# A solution to generalized learning from small training sets found in infants' repeated visual experiences of individual objects

**Frangil M. Ramirez**[1] **Elizabeth Clerkin**[2] **David J. Crandall**[1] **Linda B. Smith**[2]
fraramir@iu.edu emclerki@iu.edu djcran@iu.edu smith4@iu.edu

[1]Department of Computer Science, Luddy School of Informatics, Computing, and Engineering

[2]Department of Psychological and Brain Sciences

[1,2]Indiana University, Bloomington, IN 47405

**Abstract**

One-year-old infants rapidly form and generalize categories from idiosyncratic experiences of very few exemplars of those categories. Here we provide evidence on the statistics of infants' daily-life visual experiences for 8 object categories. Using a corpus of infant head-camera images recorded at mealtimes (87 mealtimes,14 infants), we measure the frequency of the unique instances of each category and the variability of the visual experiences within and across instances of the same category. The frequency distributions of instances for individual infants are highly skewed, containing many images of the same few objects along with fewer images of other instances. Graph theoretic measures of individual category experiences for individual children reveal a "lumpy" mix of high similarity and high variability, organized into multiple but interconnected clusters of high-similarity images. In computational experiments, we show that artificially created training sets characterized by an interconnected mix of high and low similarity support generalization to novel instances after limited training. We discuss implications for category recognition, and for learning more generally, by both humans and machines.

## 1 Introduction

The visual objects labeled by common nouns — bowl, cup, chair — are so readily recognized and their names so correctly generalized by young children that early theorists suggested that these "basic-level" categories "carve nature at its joints" (e.g., (1, 2)). This idea is at odds with contemporary understanding of visual object recognition in both human visual science and computer vision. In these literatures, object recognition is seen as a hard and not-yet-solved problem (3-5). If object categories are "given" to young perceivers, theorists of human and machine vision do not know how.

In a remarkable diary study, Mervis (6) tracked the objects seen and names heard by her one-year-old son during daily life. Her son's experiences of things the family called "duck" illustrate the overall pattern observed. Three individual objects dominated the infant's "duck" experiences: a life-like mallard toy, a soap dish, and a Donald Duck Pez dispenser. How could this child form a generalizable concept of duck given this input? Although this one child's experiences of duck instances are idiosyncratic, all infants' experiences are constrained by time and place, and therefore likely consist of many repeated experiences of a small set of specific instances. Mervis (6) worried that her child could not generalize from experiences with so few instances. However, the current experimental evidence indicates that between the ages of 6 to 12 months, infants form generalizable categories that go beyond their experienced instances. Given a heard object name and two pictures of objects on a screen, 6- to 12-month infants preferentially look at the image of the named thing even when the pictured thing is unfamiliar (3, 7-9). For a small set of categories prevalent in infant everyday life, infants generalize heard names to never seen before instances of the named category.

The theoretical consensus in both human cognition (10-12) and machine learning (13-16) is that category generalization requires transformations of the similarities of instances in the form of changes in the feature weights that capture the defining properties of category membership. The empirical consensus, from experimental studies of human learning and from machine learning, is that larger training sets with diverse instances lead to better representational spaces and therefore better generalization (17-21). This consensus does not appear to align well with infant repeated experiences of just a few instances.

The infant's training data, although small in the number of unique instances, is not small in the number of unique images. In the everyday context of active infants and a 3D world, the images received by the perceiver of any individual object will vary by small and by large amounts as a function of the spatial relation of the object to the perceiver, the lighting, and the context of the surrounding scene. Thus, an infant's repeated experiences of a single instance — for example, their own sippy cup (Figure 1A) — will create a training set containing images of the same thing that are often highly similar but sometimes very different. Our hypothesis is that the similarity structure of these early training sets – sets like those observed by Mervis (6) – are characterized by a mixture of high and low similarity images (19, 22-27). We propose that this similarity structure yields early changes to an emerging representational space that support generalization to novel instances despite the smallness of the training sets.

We used a corpus of head-camera images collected by infants aged 7 to 11 months (n = 14) over several days (28, 29). Although infants this age do not produce many words, the age range constitutes the front end of generalized receptive knowledge of the names for common categories (7-9). From the corpus, we extracted mealtime episodes, defined as continuous events in which food or dishes were present. Mealtimes are a useful context because many of the categories with the earliest learned names are present (29-31). Further, for infants, contexts involving food and dishes (real and toys) occur on average 5 times a day and in highly variable contexts (at a table, on the floor, in the car, in the sink, in a playroom, at the park, etc.). Thus, mealtimes offer the opportunity for both similarity and diversity in instances and for repeated experiences of the same instance in different visual contexts.

First, we quantify the frequency of unique instances in toddlers' visual experiences of eight early-learned categories, extending the contribution of Mervis (6). Second, using graph theoretic measures, we quantify, for each infant and each category, the similarity structure of repeated experiences of single instances and the aggregated similarity structure of all instances. Third, in machine learning experiments, we create very small artificial training sets with and without the observed similarity structure of infant daily-life experiences and assess generalization to novel instances of the trained categories.

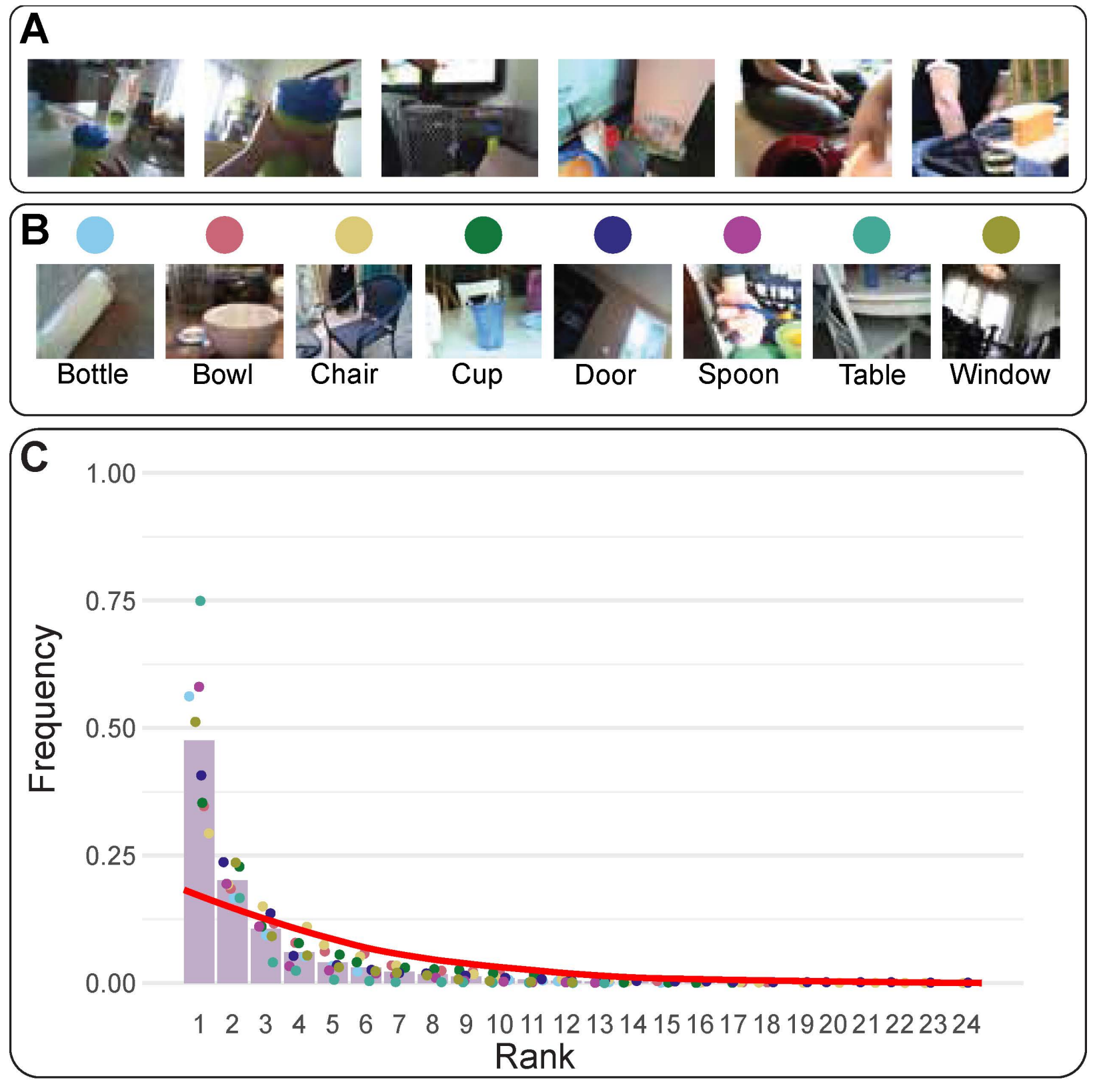

**Figure 1.** A skewed distribution of infant experiences of instances of early-learned object categories. **(A)** Head-camera images of cups showing different images of the same cup and different images of different cups. **(B)** The 8 target categories **(C)** Mean rank order frequency distribution of unique instances across all mealtimes, normalized as a proportion of all ranks for each category. For each category and subject, we count the number of occurrences of each object instance and rank the instances by frequency in descending order. The colored dots display each category, and the bars illustrate the mean across categories (20,339 datapoints in total). The (smoothed) red line illustrates the null distribution under the hypothesis that infants uniformly sample experiences from the instances present in images recorded in their home.

## 2 Results

### 2.1 Infant Visual Experiences of 8 Object Categories

The object names receptively known and generalized by 6- to 12-month-olds in controlled laboratory experiments (7-9) are all among the earliest nouns receptively understood by children according to large-scale normative studies based on parent report (30, 32), a measure that indicates older ages of acquisition than is demonstrated by controlled laboratory studies. From the corpus of 87 mealtimes, we selected 8 basic-level object categories (33) that were visually frequent in the corpus (28, 29) and among the earliest learned nouns according to those parent report norms. By those norms, the 8 target categories had a median receptive age of acquisition of 14 months (range 10-16 months, (28-30, 32)). The 8 categories include both holdable objects (bottle, bowl, cup, spoon) and larger "background" objects (chair, table, door, window). No meaningful differences were observed for the classes of holdable versus background objects in any of the following analyses. Images were sampled from the mealtime video clips at 0.2 Hz, yielding 11,549 images for analysis. Human coders determined the presence of unique instances of the 8 target categories in each image.

The collected head camera images exhibit the referential ambiguity of egocentric daily-life visual scenes. Single images typically contained instances of multiple object categories, including both the target categories (e.g., a cup and a table; Fig. S1A) as well as other categories. However, within a single image, there was often just one instance of any present target category (e.g., one cup and one table; Fig. S1A). Averaged across categories, 0.62 of images containing a given category had exactly one instance of that category (SD = 0.16; range 0.37 for chair to 0.89 for table).

Mealtimes occur as continuous episodes with the properties of the input tightly constrained by time and place. Within an individual mealtime, a single instance of any present category occurred with high frequency (Fig. S1B). The single most frequent instance of a visually present category within a mealtime constituted, on average, 0.61 of all experiences of that category within that mealtime (SD = 0.14; range 0.39 for chair to 0.86 for table; Fig. S1B). The three most frequent instances of a visually present category accounted for, on average, 0.92 of the appearances of that category within a mealtime (SD = 0.06; range 0.79 for chair to 0.99 for table). Thus, individual mealtimes present repeated experiences of just a few instances.

Across the aggregated mealtime experiences of an individual infant over several days, we observe the same pattern as reported by Mervis (6) in her diary study: very few instances dominated the across-mealtimes input of each child's categories. The most frequent instance (Fig. 1C) accounted on average for 0.48 of all experiences (range across categories from 0.29 for chair to 0.75 for table). Across all mealtimes for an infant, Ranks 1, 2, and 3 constituted on average 0.78 of all infant experiences of that category (range across categories 0.64 for chair to 0.96 for table). Thus, across multiple mealtimes, infants experience a small set of the same things with high frequency. However, the distribution of instances also includes a tail of more rarely present instances. If we assume, as the null hypothesis, that individual infants uniformly sample the instances present in images recorded in their own daily-life experiences (Fig. 1C), then the skewed distribution of observed mealtime instances differs reliably from the null hypothesis (KS test; $D = 0.33$, $p < 0.001$).

The ubiquity of skewed frequency distributions in human experience is well-known (22, 28, 34, 35). Although uniform or random samplings dominate in experimental training sets in human research and in machine learning, researchers of both human and machine learning have conjectured that skewed distributions might be beneficial to learning because they provide a combination of both high similarity and high variability (19, 22-27). However, in these discussions, repeated experiences are typically treated as verbatim copies (23, 24, 27). In the 3D world of active vision there is no exact repetition of a visual experience, as the image projected to the perceiver can vary considerably with small postural movements. Over the course of multiple encounters, the image differences for the same instance can be even greater than across some pairs of different instances, when the same cup, for example, is seen right-side up, upside down, on the floor, in the dishwasher, mostly occluded, or perhaps reflecting the purple color from a nearby screen. We ask: What is the similarity structure of these highly frequent and variable repetitions of the very same thing? How do these repeated experiences of single things influence the similarity structure of the aggregate set of frequent and infrequent instances?

As is well known in both human vision (36-38) and computer vision (39-42), similarity – and how to measure it – is complicated because different classes of stimuli (faces, objects, letters of the alphabet) have very different properties. For example, color histograms are sufficient to identify varying views of some objects and scenes in natural photographs, but not for identifying black letters of the alphabet or faces (43-49). There is also the question of whether one should measure similarity in terms of the whole image or for a cropped region around the target object. We chose to measure the similarity of whole images for 3 reasons: (i) Infants receive whole images without bounding boxes. (ii) The processes through which bounding boxes are determined (by adult human coders or computational algorithms based on adult human coders) may not be available to infant perceptual systems. The generation of a bounding box uses expected geometric content in the image, an approach challenged by the combinations of occlusions, unusual views, and atypical postures that characterize infant daily-life egocentric images (50-54). (iii) The content of whole scenes is known to play a role in object detection and category learning which may be critical to early learning (55-58). Accordingly, we measured the similarity of pairs of whole images at the raw RGB histogram level (43-45), appropriate to the images and to the fact that 7- to 11-month-old infants are novice learners who may not have acquired the higher-level features that emerge as a product of category learning. Measures at three other levels – low visual to semantic – were also conducted and show consistent patterns (Fig. S2AB). All pairwise similarity measures are computed within participant and within category.

The aggregated overall distributions of within-participant, within-category pairwise similarities reveal many highly similar pairs but also a wide range of low to very low similarity pairs (Fig. 2A). To determine the contribution of high frequency individual instances, we partitioned the overall distribution of similarity pairs for each category and infant into three groups: within Rank 1 (the most frequent instance) pairs (R1-R1), Rank 1 to others (R1-O), and all others (non-Rank 1) to each other (O-O). The distribution of R1-R1 pairs differed from the distribution of R1-O pairs (KS; $D = 0.14$, $p < 0.001$) and from the distribution of O-O pairs (KS; $D = 0.09$, $p < 0.001$), with R1-R1 pairs containing more high-similarity pairs. The mean similarity of R1-R1 pairs was also greater (Wilcoxon rank-sum test, $p < 0.001$; Welch's two-sample t-test, $p < 0.001$) than R1-O pairs and O-O pairs. For all 8 categories, R1-R1 pairs also include many highly dissimilar images (Fig. S2A). In brief, images of the very same object captured by the same infant are often highly similar but sometimes also very different.

The aggregated within-category pairwise similarities for each participant provide an estimate of the daily-life similarity structure of infant experiences. We computed the entropy of these similarities, which was less than that of the null distribution (Fig. S3B, permutation test, $n=9999$, $p < 0.001$; SI Methods). This indicates that the overall within-category predictability of similarities from one image to another is greater in infant experiences than in the null distribution that assumes a uniform sampling of available instances for the category. The category similarity statistics for each of the 14 infants show similar patterns (Figure S3C). In brief, infant visual experiences constitute a training set that contains many high similarity pairs but also many low similarity pairs, with the most frequently repeated instances contributing both high and low similarity pairs.

Network theory (59, 60) provides a way to both visualize and quantify the within-category structure. We created graph theoretic networks for each infant for each of the 8 categories (Fig. 2B). Each node represents an image and the edges between nodes represent high pairwise similarities (Methods, SI). Consistent with network theory (61-63), we report edges that represent the top 10% of measured similarities. Thus, all edges in the analyzed networks indicate relatively high similarity. High similarity is relevant to human learning because the similarity of one input to another influences the likelihood of the joint activation necessary for connection in human memory (64). Analyses using higher and lower similarity thresholds for the edges show the same patterns (Fig. S2CD). In summary, infant-by-category networks (Fig. 2B, S5-S8) reveal a pattern of densely interconnected pairs of images and sparser connections between those different clusters. We will use the non-technical term "lumpy" to refer to the observed, and not yet understood, pattern of pairwise similarities.

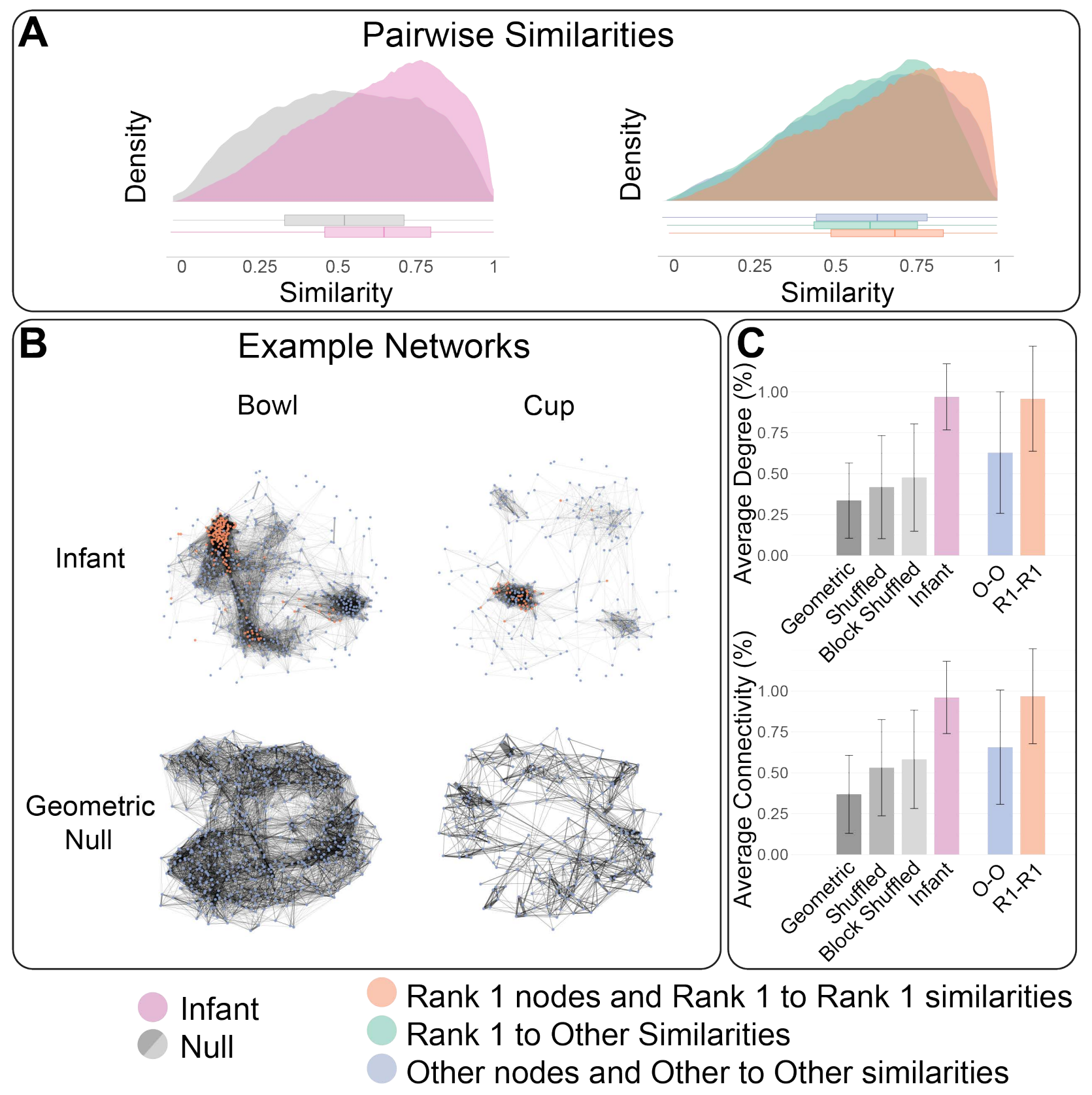

**Figure 2.** The similarity structure of infant experienced instances. Similarity is measured as RGB histogram correlation $c_{i,j}$ between pairs of images $(i,j)$. **(A)** *Left*: Pairwise similarity density aggregated across the eight target categories (1,188,150 datapoints for the observed infant distribution and 97,118 for the null distribution). The box plots illustrate the 0.25, 0.5 (median), and 0.75 quartiles. *Right*: Observed pairwise similarity density partitioned into three groups of pairs: within Rank 1 pairs (R1-R1), Rank 1 to others (R1-O), and all others (non-Rank 1) to each other (O-O). There are 1,037,730 datapoints in total. See Methods for explanation of differing numbers of data points. **(B)** *Top:* Example of two observed networks for one infant. *Bottom:* The corresponding randomly generated geometric null networks. Each node represents one image, and the edges indicate pairwise similarities in the top 10% of measured similarities. **(C)** *Top*: average degree (LMM; Infant vs. Nulls, $p<0.001$, $n=94$ networks per bar, $N=376$ networks in total). *Bottom*: average connectivity (LMM; Infant vs. Nulls, $p<0.001$, $n=94$ networks per bar, $N=376$ networks in total). The metrics are further partitioned into Rank 1 nodes (orange) and Other nodes (blue); LMM, Rank 1 vs. Other, $p<0.05$, $N=120$ networks in total. The error bars illustrate corpus-level standard deviation.

Because the infant-by-category data sets are small and because time and space constrain the possible network patterns that could occur independent of category membership and context, we created three sets of null graphs with different constraints. The first, Shuffled Labels, is the least constrained and constitutes a common approach to forming a null hypothesis in graph theory. We shuffled the infant-category labels for the images. If the similarity properties and spatial relations within the images themselves create the observed patterns, unrelated to category membership, they should be evident in this null. The second null, Blocks of Shuffled Labels, adds the constraint of temporal proximity of images to each other, since close-in-time images are likely to be similar. We shuffled the labels for contiguous blocks of 6 images (~30 seconds, Methods, SI). We include the data for smaller and larger block sizes of label shuffling, which show consistent patterns, in the SI (Methods, Fig. S4, Tab. S1). If blocks of close-in-time images create the lumpy pattern, it should be evident in this null. The third null, a Geometric null, is based simply on the geometry of a bounded similarity space. For each infant and category, the observed number of nodes were positioned uniformly at random within the unit square and then nodes were connected that fell within the top 10% similarity range in the corresponding infant graph (65). If simply thresholding a bounded similarity space at a given radius creates the observed patterns, independent of any image statistics or semantic structure, it should be evident in this null.

The key theoretical question to be answered is the connectivity of the training instances to each other. A first index is the number of images (nodes) that are not connected to any other image, that is, isolated nodes. The observed infant graphs showed fewer isolated nodes than all three null versions (Linear Mixed-effects Model [LMM], $p < 0.05$; Methods). The proportion of isolated nodes was on average 0.20 (SD = 0.12), 0.26 (SD = 0.28), 0.50 (SD = 0.24), and 0.43 (SD = 0.24) for the Infant and Geometric, Shuffled Labels, and Blocks of Shuffled Labels nulls, respectively. The two Shuffled nulls, which are less constrained than the Geometric null, have many isolated nodes relative to the Infant data.

We used two graph-theoretic measures to quantify the properties of the observed networks and the three classes of null networks (Fig. 2C). The first measure, average degree, computes the number of edges per node and then averages across nodes (66). Isolated nodes contribute a degree of zero (Methods). Graphs with higher average degree have more direct connections among the individual nodes. Overall, the observed infant graphs showed higher average degrees than all three null versions (LMM, $p < 0.001$). The three nulls did not differ reliably from each other (SI). These comparisons suggest that the high average degrees observed within infant-experienced instances do not derive solely from the set of images and their temporal properties, but reflect image properties relevant to target category membership. The second measure, average node connectivity (67), measures the mean number of node-disjoint paths between all connected image pairs (and does not include isolated nodes). Higher values indicate greater robustness within the connected network components: more images (nodes) would need to be removed before pairs of images would become disconnected (no remaining high similarity path to other images). The average node connectivity is greater for the infant data than each of the nulls (LMM, $p < 0.001$), and higher for Blocks of Shuffled Labels than the Geometric null (LMM, $p < 0.05$) with no other reliable differences between the nulls. A conclusion that the Blocks of Shuffled Labels null has better connectivity than the Geometric null is not warranted because it has more isolated nodes. In brief, the observed Infant networks differ reliably from each of the three nulls on all measures, suggesting that children's varied experiences of just a few high frequency instances and other less frequent instances create a data set with robust connections among the individually experienced instances.

The high frequency experiences of very few instances appear to play a role in creating the lumpy pattern of pairwise similarities. Within the infant data, the average degrees for R1-R1 nodes were greater than for R1-O nodes (LMM, $p < 0.001$) and O-O nodes (LMM, $p < 0.05$). Likewise, the average node connectivity is greater for R1-R1 nodes than for R1-O nodes (LMM, $p < 0.001$) and O-O nodes (LMM, $p < 0.05$) in the observed data (Fig. 2C).

In summary, infant visual experiences of instances of 8 early learned object categories over multiple mealtimes and days are characterized by a skewed frequency distribution in which a few instances account for more than half of all experiences of that category. The repeated experiences of the same instances are themselves visually variable, contributing both high and low similarity images of the target object. If just high similarity – easy-to-connect – pairs are considered for a participant and category, one sees a lumpy distribution: multiple dense regions of highly similar images that are connected to each other through sparser pairs of high similarity images. The new and open question is whether this observed structure in infant experiences is part of the explanation of how infants, from their limited and idiosyncratic experiences, manage to generalize early categories to novel instances.

### 2.2 Small, lumpy datasets and machine learning

The present findings concern a period in human development, prior to the first birthday, in which infants generalize the names of categories to new instances (7-9). These generalizations suggest that infants know something about the category-specific properties that determine category membership (1, 2). Category knowledge is by no means complete at this point; generalization errors relative to adult categories do occur, but appropriate generalization is more common than not (3, 6, 68). To provide evidence for the plausibility of the idea that "lumpy" training sets contribute to a representation space that yields appropriate generalization despite limited training experiences, we artificially engineered extremely small training sets that either did or did not have the observed properties of the infant data. We trained three different models from scratch (no pretraining) and then asked if the models could generalize the trained categories to novel instances of those same categories. Our goal is not to model infant category formation nor to beat machine learning benchmarks but to provide insight to the possible role of lumpy similarity structures in supporting generalization from very small training data. We compared the engineered Lumpy training sets to a Uniform sampling of instances that yields a more diverse training set. Uniform training sets are generally assumed better for training than over-sampling a few instances (19, 69).

To create the experimental training conditions (Lumpy and Uniform) and Generalization Test sets, we used two publicly available data sets consisting of visual categories with varied images of the same entities (Fig. 3A). ShapeNet (70) contains 3D models of individual object instances, allowing us to render varied 2D views of the same unique object, and PetFace (71) contains multiple images of the faces of individual animals, allowing a training set with multiple views of the very same animal (e.g., multiple views of Rover the dog). Although there are in principle multiple ways to engineer lumpiness in a training set, we followed the path that created the infant data, oversampling variable views of a small set of instances relative to other more rarely presented instances within each category (Methods). Thus, in the Lumpy training condition, the training set has both a highly skewed frequency distribution of instances and variable views of all instances. For the Uniform condition, we uniformly sampled variable views from all instances, and thus this training set includes fewer varied views of the same instances and no skew in the instance frequency distribution. The frequency distribution of categories (not instances) is uniform in both conditions. The training sets are very small: a total of 1,500 unique training images for PetFace (250 images × 6 categories) in both the Lumpy and Uniform conditions and a total of 3,600 unique images for ShapeNet (~600 images × 6 categories) in both the Lumpy and Uniform conditions. The pairwise similarity distributions within each training category were measured by the same graph theoretic properties used in the infant analyses, and showed that for both the engineered ShapeNet and PetFace training sets, the Lumpy training sets have more high similarity direct connections and greater node connectivity than the Uniform training sets (Tab. S2, S3).

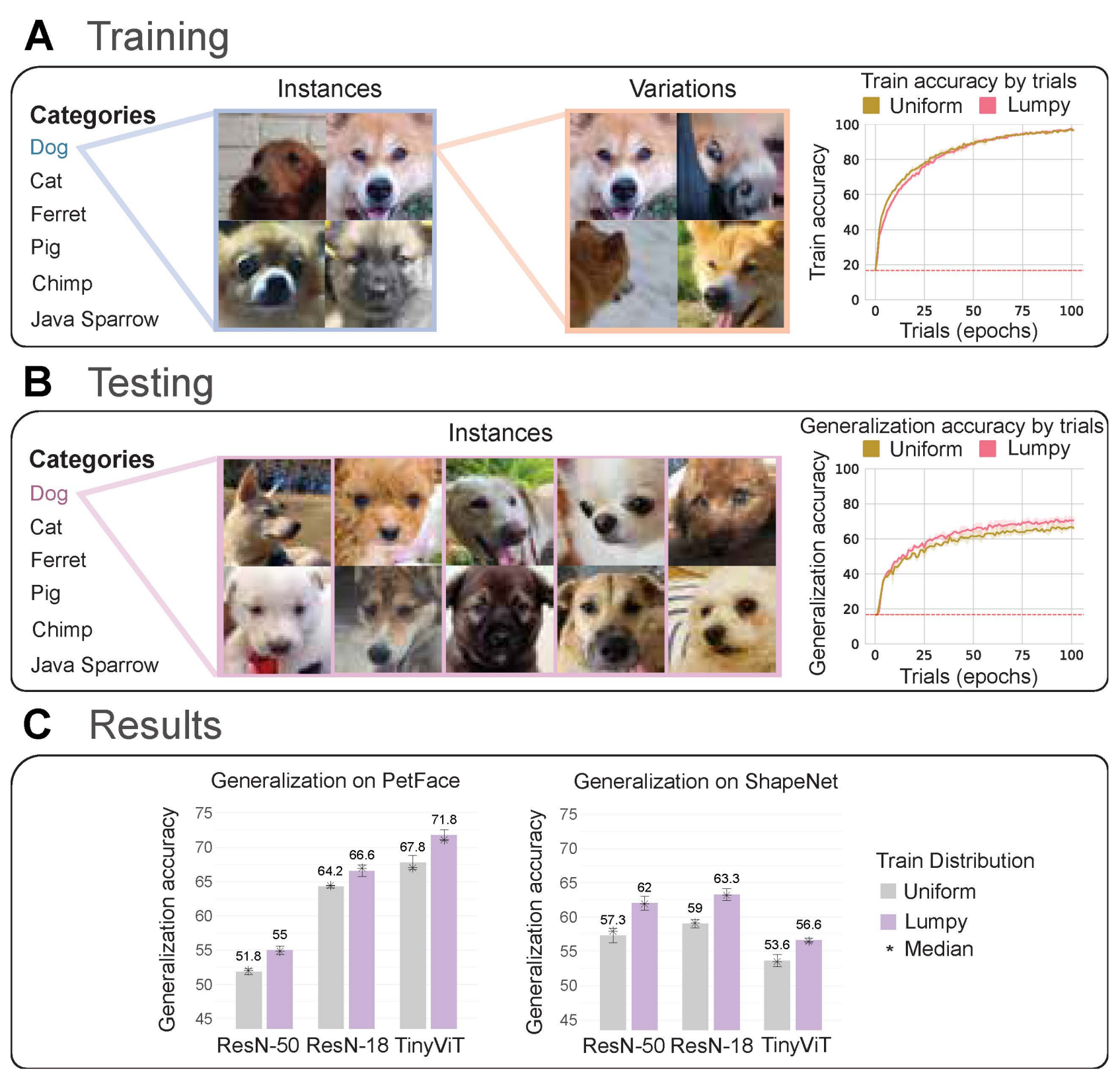

**Figure 3.** Experimental setup and results of machine learning models. **(A)** Training setup. *Left*: For each category, we generate Lumpy training sets by oversampling variable views of individual instances, and Uniform training sets by uniformly sampling experiences from available instances. *Right*: Models trained on Lumpy and Uniform sets reach near-perfect training accuracy. The bars display the mean and 95% CI over three runs of a TinyViT model on each set (6 total). The dashed line illustrates chance performance. **(B)** Generalization test setup. *Left*: To measure generalization, we build a test set with novel instances, e.g., individual dogs, not seen during training. The same test set is used to measure generalization of the Lumpy and Uniform training sets. *Right*: Models trained on Lumpy sets achieve better generalization to novel instances than their corresponding Uniform training sets. The bars display the mean and 95% CI over three runs of a TinyViT model on each set (6 total). The dashed line illustrates chance performance. **(C)** Generalization test results on both PetFace and ShapeNet datasets. Training on Lumpy sets yields better generalization to novel instances than training on Uniform sets in all cases. ResNet-50 and ResNet-18 (abbreviated as ResN.) are each trained 8 times on ShapeNet, and 6 and 3 times on PetFace, respectively, due to lower variance. TinyViT models are trained 3 times on each dataset. LMM, Lumpy vs. Uniform, $p<0.001$, $N=62$ runs in total. The error bars illustrate standard error.

The Generalization Tests measure generalization of the trained categories to novel instances of those categories. For example, if the faces used to train the category “dog” belonged to individual dogs A (“Abby”), B (“Butch”), and C (“Champ”), then the test faces for the “dog” category would be from other individual dogs D, E, and F with none of their images included in the training set. The test sets consisted of 1,500 PetFace and 3,600 ShapeNet images of variable views of novel instances not seen during training (see Methods, SI). Within the ShapeNet and PetFace instantiations of the experiment, the same Generalization Test set was used for all training conditions.

We trained three different models, two ResNet Convolutional Neural Networks (CNNs) with 18 and 50 layers (72) and a TinyViT (a very small Vision Transformer, (73)). We used supervised learning from scratch because experimental measures of infant category generalization prior to the first birthday are based on heard names (7-9). We acknowledge that infant learning is likely based on semi- and multi-supervised signals of category membership that include labeling from heard names, co-occurrence information (e.g., a cup beside a spoon), and actions (74, 75). Here we use supervised learning with clear category labels to provide a strong test of the idea that very small training sets with a lumpy similarity structure support generalization from exposure to a small number of instances.

For all instantiations of the experiment, across all models and data sets, the Lumpy training sets yielded better generalization to novel instances in the test set than did the corresponding Uniform training sets (Fig. 3C). A linear mixed-effects model showed a significant main effect for the similarity structure distribution (LMM, $D = 3.83$, 95% CI [2.8, 4.8], $t(55) = 7.42$, $p < 0.001$; Lumpy vs. Uniform) with no interactions with model or dataset. The overall advantages of the Lumpy versus Uniform training sets are small but we believe meaningful. First, they hold across all instantiations of the experiment and thus are not dependent on the specific stimuli or model architecture. Second, training with a small data set in which a very few instances dominate is no worse than training with a uniform set of training instances which is generally expected to be the optimal distribution for generalization. Third, generalization to novel instances in these models depends on a representational space in which novel instances are embedded near prior learned instances of the same category. Thus, the machine learning results indicate that training sets in which a few (varied experiences of) unique instances dominate can shape the representational space in the direction of appropriate category generalization.

We began this study with the hypothesis that infant exposures to instances of early learned categories would, like those of Mervis (6)'s child, be dominated by a few idiosyncratic specific things. This seemed likely because of the constraints of the continuity of time and space on real world experiences. We asked how, if this was so, an infant could possibly recognize a novel instance of those categories, which experimental data indicates that they do (7-9). The present findings show that Mervis (6)'s observations are not specific to one child. They also show that real world experiences of a just a few instances present a pattern of mixed high and low similarities in which there is high similarity connectivity across pairs of input despite overall diversity. The machine learning experiments show that this Lumpy structure, the product of repeated but varied experiences of a few individual things, supports generalization to novel instances. The lumpiness in the infant experiences and generalization by the machine learning models are not yet directly connected to infant category generalization. Experiments with infants in which training sets are manipulated and category generalization is tested is a critical next step.

## 3 Discussion

Human visual experiences provide relevant varied information about objects at two levels of granularity. Theories of perception traditionally treated the two levels as two distinct computational problems (76-78). The first, object constancy, concerns the perception of an individual object as the same thing with the same stable essential properties despite the variability in the input. The second, categorization, concerns the ability to recognize never-seen-before things as members of categories. The two levels of variability are mixed in daily-life experiences and in daily-life recognition tasks (my sippy cup, my mother's cup, a new cup). The field does not have a unified understanding of the interplay of variable experiences of a single thing and experiences of different instances of the same kind in category formation. The present findings show, first, that infant everyday experiences of object categories are dominated by variable experiences of single things, a fact that strongly suggests that the field needs to consider the role of this input in category formation. The observed similarity statistics of the combined variability of images from the same and different things reveals a similarity structure that has not been studied in human or machine learning but that provides testable hypotheses about how generalizable object categories may be acquired from experiences dominated by variable images of just a few idiosyncratic instances. In so doing, the findings make a clear case for increased research on the interplay of the varied input of single things and across individual things that are the same kind — behaviorally, neurally, and computationally (3, 4, 76, 79). The findings also highlight the value of determining the statistics of the daily-life experiences on which human perceptual and cognitive development depends.

Theories of category formation and generalization in psychology, neuroscience, cognitive science, and machine learning (10, 12, 80-83) conceptualize learning as the warping of a similarity (embedding) space through increasing and decreasing weights on features to achieve compression within categories and expansion across boundaries between categories. Infants' ability to generalize familiar categories suggests that their limited (and idiosyncratic) experiences change the similarity space in ways that support generalization. The implication is that the lumpiness of the similarity structure of the input – observed in infant egocentric experiences and manipulated in the machine learning experiments – shapes the emerging representational space in ways that speed up learning about rarer and newly encountered instances. This hypothesis needs to be directly tested.

There is precedence for this idea in the classic computer vision literature on basis functions (78). Poggio and Edelman (78) developed an approach based on the theory of approximation of multivariate functions that learns from a small set of perspective views a function that maps any viewpoint to a standard view. The key insight is that the set of views form a low-dimensional manifold, and a small number of stored views suffice as radial basis functions to interpolate across the whole manifold. These ideas were later extended to learning multiple categories (84). Although their approach does not mathematically or computationally align with the machine learning models used in our analyses, it points to the larger idea that the right small slice of data at the start of learning, with the right statistical structure, may set up an embedding space that, although not in its final form, is warped in the right direction. Our idea that varied exposures to a few individual objects may be sufficient for category learning has also been raised in several recent computer vision studies (85, 86). There are also precedents in theories of phonological development in infant representation of speech sounds. Empirical results and well-developed theories reveal that early-learned speech sounds create what are called "perceptual magnet effects" that form an initial shaping of the representational space biased to the speech sounds in the language being learned (87-89). The intersections of these ideas are worth pursuing if we are to understand how a small set of common categories are rapidly formed and generalized by infants in the period before their first birthday. The overarching hypothesis is this: If a developing representational space "knows" how the images (and contexts) of one sippy cup can vary, then the representation space may also know something about how the images of a distinct but same kind of thing can also vary (Fig. 4A). The distinct clusters of high similarity images may warp the representational space in ways that interconnect the varied of experiences of individual and different things to support generalization (Fig. 4B).

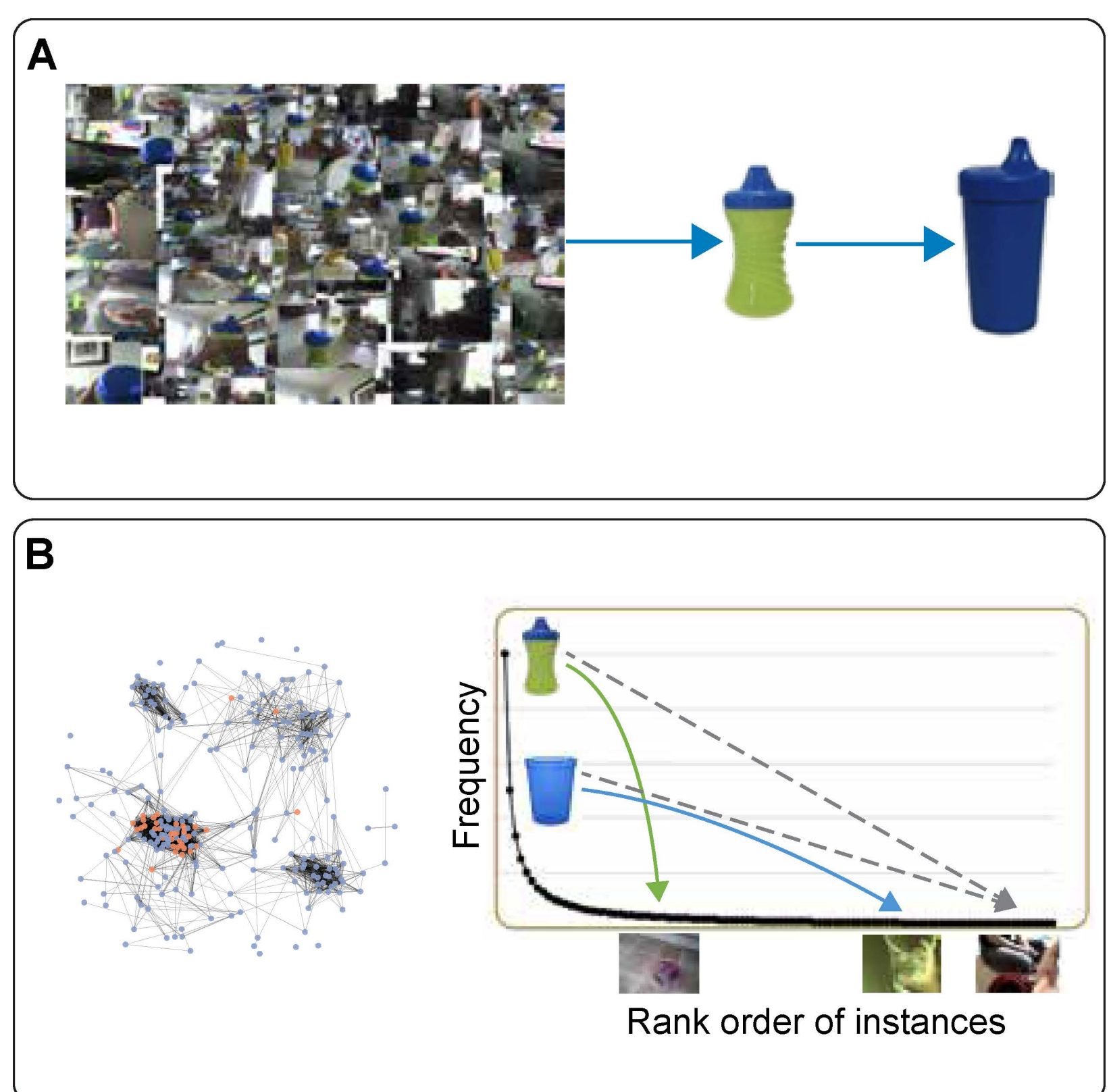

**Figure 4.** A hypothesis about the relation between object constancy and category generalization from few experienced instances. **(A)** Repeated experiences of a single object yield highly variable images of that object that share the stable properties of the object. Thus, repeated experiences with one object may support generalization to other objects. **(B)** The lumpy similarity structure generated by a mixture of repeated experiences with a few unique instances and rarer experiences with other instances may support the efficient formation of generalizable representations by connecting similar images of different things that then lead to latent knowledge (grey arrows) of the defining properties.

### 3.1 Limitations and open questions

Is the role of varied experiences of a few individual instances along with a long tail of more rarely encountered instances specifically advantageous to the earliest stages of category formation or a general principle for learning at any stage? A large literature on human learning suggests that while increased variability is generally advantageous, consistency in the input is often beneficial to novice learners (19, 23). Category learning by 6- to 12-month-old infants is necessarily small-scale given the constraints of time, space, and their limited cognitive and motor skills. Thus, the statistics of the input may change as infant abilities advance, opening doors to more diverse experiences of instances (22), and given the prior formation of earlier categories, learning in older infants and adults could generally benefit from greater diversity. This idea – early training on a small set of instances with the right structure and later training on more diverse instances – might also benefit machine learning (19, 25, 27, 90, 91). However, the present findings are not straightforwardly about high or low variability in the training sets but about the kind of variability – variations (small and large) of experiences of the same instance and variation across different instances. The specific visual properties that remain constant and that differ between variable images of the same instance and variable images of different instances need to be determined along with their role in category learning.

Several further open questions about lumpiness suggest caution in interpretations. The structure of the infant data suggests that lumpy similarity structures emerge from a skewed frequency distribution of instances and from variations in the images of those instances. The contributions of the Zipfian-like frequency distributions and variations of unique instances need to be disentangled. It is also possible that the observed lumpiness of whole-image similarity is not itself the significant factor in the training data but co-varies with a yet to be determined critical property of the infant data sets. Future work in experimental infant category learning is needed to determine the roles of object similarity and diversity versus context (background) similarity and diversity in category formation and generalization. Both are likely factors (58) but currently undetermined. A related open issue concerns the relation of lumpiness as described here and augmentation in machine learning. The use of dataset augmentation in computer vision was originally introduced to increase the size of training sets by creating multiple synthetic transformations (e.g., rotations, partial occlusions, cropping, color shifts (92)) of original training images, but has since been found to increase generalization even when training set size is controlled (93). The beneficial effects of this tactic are not yet fully explained (16, 94-97) but may provide insight into how to engineer training materials with lumpy similarity structures that support rapid generalization by humans and machines. One possible implication is that the value of lumpy training sets is more general than the time- and place-constrained experiences of infants, that it is about the optimal statistical structure of any training distribution at any scale for datasets with many category-relevant stimulus features (23).

One might ask if the observed properties of the infant experiences in this study are specific to mealtimes of infants prior to their first birthday in a specific culture or context. We cannot of course answer this question with any certainty. Extant studies of the daily-life experiences of children developing across the world's diverse human communities indicate both deep similarities as well as significant differences (98-100). However, Zipfian distributions have been shown to characterize a wide variety of self-generated human behavior, including where we look, what we touch, and where we go, and do so across multiple temporal scales (35, 101-103). The visual input to freely moving perceivers in the physical world are inherently and continuously variable. Thus, although the objects encountered by infants in different daily-life activities and in different human communities may vary in important ways, it is likely that the frequency distribution of encountered individual things and their visual variability is a shared characteristic of human experience. A critical open question is the generality of the observed similarity structure across different categories and domains of human self-generated experiences.

Although cautions are required in making strong conclusions, the findings suggest both new insights and new research paths to understanding how human infants form early object categories from relatively limited experiences with instances of those categories. When the efficiency of human learning exceeds that of known learning mechanisms, theorists (104, 105) often postulate the existence of "given" intrinsic knowledge in the form of "inductive biases." Although the present findings raise many open questions, the results suggest that the solution to efficient learning may also lie in the statistics of the natural experiences of infants and children, the training data foundational to human intelligence.

# 4 Materials and Methods

## 4.1 Infant Visual Experiences of 8 Object Categories

**Data collection.** The corpus of egocentric images during mealtime events is a portion of a larger data set of head-camera videos collected by infants in their homes, $n$ = 100, from 3 weeks to 24 months old (106). The camera recorded at 30 Hz and it was attached to a headband that held the camera low on the toddlers' forehead directly above the nose. Infants and toddlers tend to orient eyes and heads in the same direction (bringing them into alignment within 500 msec) when shifting attention (107-110). Studies using head-mounted eye-tracking demonstrate that 80% of gaze by infants and children falls within 20 degrees of the center of the head-camera image (111, 112). Parents were taught to place and operate the head cameras and were asked to collect 6 to 8 hours of daily-life experiences; no experimenters were present in the home during data collection. Parents were told that the study was about early visual experiences and that all visual experiences were relevant. The subcorpus of 87 mealtimes (mean duration of a mealtime = 11.22 min., SD = 11.87) was extracted. Analyses of the objects in view and the object names heard using the mealtime corpus were reported in two prior studies (28, 29).

**Identification of instances in images.** Adults were paid to label the objects in each image (see (28)). From this coding we selected the 8 target objects that had high frequency in the mealtime corpus and whose names were normatively among the first 100 nouns learned by children learning American English (30). The number of instances in an image and the identity of the individual instances were coded by trained adults in the laboratory. The unit of analysis is an object occurrence, meaning that an image containing more than one target category is counted as a datapoint for each of those categories.

Computations for all reported measures and figures on the infant data are in the Appendix SI.

## 4.2 Small, lumpy datasets and machine learning

Training and testing data sets were created from a subset of the publicly-available ShapeNet (70) and PetFace datasets (71). We train three computer vision models on paired Lumpy and Uniform datasets and measure generalization on the same test sets. Although there are likely multiple ways to engineer lumpiness in a training set, we followed the path that created the infant data, oversampling some instances relative to others. We explain the data generation process in more detail next.

**ShapeNet dataset.** This dataset contains three-dimensional models of individual objects (70), which we used to generate two-dimensional images using rendering software developed by us. Six object categories were used: airplane, bed, bottle, car, chair, and camera. Eight object instances were used for each category, for a total of 48 instances. We generated a set of views for each instance by rendering it at random 3D orientations and saving each view as an image. We generated 600 views per object instance. To do this, we sampled a set of random orientations (Euler angles) exactly once from a uniform distribution in the range $[-180, 180]$ and used the same set of angles for all objects. Although this simple sampling approach does not produce a perfectly uniform distribution over the rotation group SO(3), it provides a broad and approximately isotropic coverage.

To generate the training sets of 3,600 images, we subsampled a subset of the images which yielded approximately 600 images for each of the six categories. To measure generalization performance, two of the eight instances per category were excluded from training and only presented to the model at test time. That is, if the category “car” is trained using car instances A, B, C, D, E, and F, then the test sets will contain novel car instances G and H. For the test sets, we selected exactly 600 images per category (300 images × 2 instances each), for a total of 3,600 images. For fairness, the exact same test images were used to evaluate all training sets, and the train and test sets are composed of 3,600 images each (1:1 ratio).

To generate the Lumpy training set, we oversampled views from a randomly-chosen training instance per category. Specifically, we sampled such that 35% of the training images for each category corresponded to the Rank 1 instance, to match our observation that the infant who contributed the most data had on average 35% Rank 1 instances across all categories. For the Uniform sets, we sampled images uniformly from all six training instances, yielding a proportion of ~16.67% (1/6) images each. The similarity measures are presented in Tab. S2.

**PetFace dataset.** This dataset contains images of individual animals’ faces (71). We again used six categories: dog, cat, pig, java sparrow, chimp, and ferret. Here, categories tend to have hundreds or even thousands of instances (individual animals), but only a handful of images for each. As each instance can contribute fewer images, we used 250 images per category. To generate Lumpy training sets, we followed the same procedure of oversampling images from a Rank 1 instance, which is simply defined to be the instance that contributes the most images of a category in the original PetFace dataset. To generate additional variable views of Rank 1 instances, we horizontally flipped (i.e., mirrored) its images. That is, for example, if a rabbit is facing right, it will face left in its corresponding mirrored image. This technique is often used in dataset augmentation in computer vision to increase overall dataset size, while here we used it to generate variable views of individual instances. We uniformly sample the remaining images from non-Rank-1 instances to obtain 250 training images per category. Note the lumpy sets thus contain fewer than 250 instances per category (mean across categories = 191.5, SD = 46.4). To generate the Uniform sets, we selected 250 instances per category and thus 1 image each. The test sets contained 250 images for each category of instances never seen during training. For example, if the faces used to train the category “dog” belonged to individual dogs A (“Abby”), B (“Butch”), and C (“Champ”), then the test faces for the “dog” category would be from other individual dogs, D, E, and F (Fig. 3AB). The similarity measures are presented in Tab. S3.

**Test alternatives and chance performance.** The number of alternatives was the number of categories used (six). That is, every test image could be classified as any of the six trained categories, yielding a chance-level performance of ~16.67% (1/6).

**Machine learning models.** We used both Convolutional Neural Networks (CNNs) and Vision Transformers (ViTs). More specifically, we used three models: a ResNet-18 CNN, a ResNet-50 CNN (72), and a TinyViT (73). The two ResNet CNNs were implemented in Pytorch (113) and TinyViT was implemented with timm (114); PyTorch and timm are are two popular open-source deep learning libraries. All models were trained using a cross-entropy loss and an SGD momentum optimizer on an NVIDIA Titan X GPU for the ResNets and an NVIDIA L40S GPU for TinyViT. Hyperparameters were defined initially and kept fixed throughout the experiments: learning rate =

$10^{-3}$, momentum = 0.9, weight decay = $10^{-3}$, batch size = 100. The learning rate was decreased once by 10% at epoch 30 out of 50. TinyViT models trained on ShapeNet used a learning rate of $10^{-4}$ to mitigate overfitting, and they were trained up to 100 epochs on PetFace as they continued improving past 50 epochs. ResNet-50 and ResNet-18 models were each trained from scratch eight times on ShapeNet, and six and three times on PetFace, respectively, due to lower variance. The test accuracy of all runs is presented in Figure 3D. TinyViT models were trained from scratch three times on each dataset, as we observed a lower variance compared to CNNs, and the test accuracy of all runs is also presented in Figure 3D.

**Input dataset statistics.** We used the same similarity measures as in the infant data. The graph networks used edges within the top 5% similarity due to a greater number of pairs in the training sets than in the observed infant data. Please see Fig. S2CD for the negligible effect of changing the threshold in the results, albeit at a higher computational cost for more relaxed thresholds. When using linear mixed-effects models to test significance, the similarity structure (Lumpy vs. Uniform) was treated as a main effect and the model instances (machine learning model × dataset combination) were treated as the random variable in the data analyses. All other measures follow the methods used in the infant data analyses (SI Methods).

## 5 Data & code availability

The subject recruiting, consenting, and data collection were approved by the Indiana University IRB. The parents of the children were contacted through an opt-in database of families administered by the Department of Psychological and Brain Sciences of Indiana University. The protocol for database and recruiting through community outreach events was conducted under Indiana University IRB protocol #809000028. The experimental and consenting procedure was conducted under Indiana University IRB protocol #1505862312. Parents were told the purpose and nature of the study; if interested, they were invited to the laboratory and the procedure was explained in detail. They were told that that they could stop participation at any time for any reason. They were told they could review the collected video and we would delete any or all of it if they desired. They were told that their privacy would be protected and that no videos or frames with identifiable faces would be used in publications or presentations without prior permission for any particular image. They were left alone to decide consent as required by human subject consenting procedures. If they agreed to participate, they were trained to put the head camera on their infant, given instructions, and took the head-cameras home. The experimenter went to the home at the end of data collection, debriefed the parent, and reviewed video with the parent and deleted any head camera videos requested by the parent.

Because the head-camera video was collected without constraints, the images contain faces of unconsented individuals as well as information through which specific families could be identified. Consequently, the sharing of the raw videos is not ethical and not allowed under the protocol. We will deposit all videos with masked face regions where applicable and raw videos that contain no identifiable faces to the Databrary repository at https://databrary.org/ available to researchers with IRB approval from their own institutions. The coding of head-camera images for the objects present in the image, their identity by participant number and category, and the pairwise similarity measures that do not contain identifiable information or faces will also be deposited to the Databrary repository. The datasheets of number and identity of instances as well the code for all analyses will be available at an Open Science Framework (OSF) repository (here).

## Acknowledgments

We thank Rick Betzel and Santo Fortunato for their helpful comments regarding the graph-theoretic analyses. The research was supported by NIH-NICHD award 5R01HD104624 and NIH-NEI award 1R01EY032897 to L.B.S. and D.J.C.

## References

1. D. Gentner, Why Nouns Are Learned before Verbs: Linguistic Relativity Versus Natural Partitioning. *BBN report ; no. 4854. Center for the Study of Reading Technical Report ; no. 257.* (1982).
2. E. Rosch, C. B. Mervis, W. D. Gray, D. M. Johnson, P. Boyesbraem, Basic Objects in Natural Categories. *Cognitive Psychol* **8**, 382–439 (1976).
3. V. Ayzenberg, M. Behrmann, The Dorsal Visual Pathway Represents Object-Centered Spatial Relations for Object Recognition. *J Neurosci* **42**, 4693–4710 (2022).
4. V. Ayzenberg, M. Behrmann, Development of visual object recognition. *Nat Rev Psychol* **3**, 123–137 (2024).
5. N. Pinto, D. D. Cox, J. J. DiCarlo, Why is real-world visual object recognition hard? *PLoS Comput Biol* **4**, e27 (2008).
6. C. B. Mervis, "Child-basic object categories and early lexical development" in Concepts and Conceptual Development: Ecological and Intellectual Factors in Categorization*,* U. Neisser, Ed. (Cambridge University Press, Cambridge, 1987), chap. 201-233.
7. E. Bergelson, D. Swingley, At 6–9 months, human infants know the meanings of many common nouns. *Proceedings of the National Academy of Sciences* **109**, 3253–3258 (2012).
8. J. Campbell, D. G. Hall, The scope of infants? early object word extensions. *Cognition* **228** (2022).
9. H. Garrison, G. Baudet, E. Breitfeld, A. Aberman, E. Bergelson, Familiarity plays a small role in noun comprehension at 12-18 months. *Infancy* **25**, 458–477 (2020).
10. R. M. Nosofsky, Attention, Similarity, and the Identification-Categorization Relationship. *J Exp Psychol Gen* **115**, 39–57 (1986).
11. R. N. Shepard, Stimulus and Response Generalization: A Stochastic Model Relating Generalization to Distance in Psychological Space. *Psychometrika* **22**, 325–345 (1957).
12. R. N. Shepard, Toward a Universal Law of Generalization for Psychological Science. *Science* **237**, 1317–1323 (1987).
13. S. Edelman, Representation is representation of similarities. *Behav Brain Sci* **21**, 449–+ (1998).
14. R. Hadsell, S. Chopra, Y. LeCun (2006) Dimensionality Reduction by Learning an Invariant Mapping. in *IEEE/CVF Conference on Computer Vision and Pattern Recognition (CVPR)*, pp 1735–1742.
15. P. Khosla *et al.*, Supervised Contrastive Learning. *Advances in Neural Information Processing Systems 33, NeurIPS 2020* **33** (2020).
16. A. Krizhevsky, I. Sutskever, G. E. Hinton, ImageNet Classification with Deep Convolutional Neural Networks. *Commun Acm* **60**, 84–90 (2017).
17. Y. Bahri, E. Dyer, J. Kaplan, J. H. Lee, U. Sharma, Explaining neural scaling laws. *P Natl Acad Sci USA* **121** (2024).
18. J. Kaplan *et al.*, Scaling Laws for Neural Language Models. arXiv [Preprint] (2020). http://dx.doi.org/https://doi.org/10.48550/arXiv.2001.08361 (accessed September 2025).
19. L. Raviv, G. Lupyan, S. C. Green, How variability shapes learning and generalization. *Trends in Cognitive Sciences* **26**, 462–483 (2022).
20. C. Sun, A. Shrivastava, S. Singh, A. Gupta (2017) Revisiting Unreasonable Effectiveness of Data in Deep Learning Era. in *2017 IEEE/CVF International Conference on Computer Vision (ICCV)*, pp 843–852.
21. R. Taori *et al.*, Measuring Robustness to Natural Distribution Shifts in Image Classification. *Advances in Neural Information Processing Systems 33, NeurIPS 2020* **33** (2020).

22. L. B. Smith, S. Jayaraman, E. Clerkin, C. Yu, The Developing Infant Creates a Curriculum for Statistical Learning. *Trends in Cognitive Sciences* **22**, 325–336 (2018).
23. P. F. Carvalho, R. L. Goldstone, Putting category learning in order: Category structure and temporal arrangement affect the benefit of interleaved over blocked study. *Mem Cognition* **42**, 481–495 (2014).
24. S. C. Y. Chan *et al.*, Data Distributional Properties Drive Emergent In-Context Learning in Transformers. *Adv Neur In* **35** (2022).
25. Y. J. Lee, K. Grauman (2011) Learning the Easy Things First: Self-Paced Visual Category Discovery. in *IEEE/CVF Conference on Computer Vision and Pattern Recognition (CVPR)*, pp 1721–1728.
26. R. Salakhutdinov, A. Torralba, J. Tenenbaum (2011) Learning to share visual appearance for multiclass object detection. in *IEEE/CVF Conference on Computer Vision and Pattern Recognition (CVPR)*, pp 1481–1488.
27. N. Zucchet, F. D'Angelo, A. K. Lampinen, S. C. Y. Chan (2025) The emergence of sparse attention: impact of data distribution and benefits of repetition. in *Advances in Neural Information Processing Systems (NeurIPS 2025)*.
28. E. M. Clerkin, E. Hart, J. M. Rehg, C. Yu, L. B. Smith, Real-world visual statistics and infants' first-learned object names. *Philosophical Transactions of the Royal Society B: Biological Sciences* **372** (2017).
29. E. M. Clerkin, L. B. Smith, Real-world statistics at two timescales and a mechanism for infant learning of object names. *Proceedings of the National Academy of Sciences* **119** (2022).
30. M. C. Frank, M. Braginsky, D. Yurovsky, V. A. Marchman, Wordbank: an open repository for developmental vocabulary data. *Journal of Child Language* **44**, 677–694 (2016).
31. C. S. Tamis-LeMonda, Y. Kuchirko, R. Luo, K. Escobar, M. H. Bornstein, Power in methods: language to infants in structured and naturalistic contexts. *Developmental Sci* **20** (2017).
32. L. Fenson *et al.*, Variability in Early Communicative Development. *Monogr Soc Res Child* **59**, R5–+ (1994).
33. E. Rosch, "Principles of categorization" in Cognition and categorization, E. Rosch, B. B. Lloyd, Eds. (Lawrence Erlbaum Associates, 1978), pp. 27–48.
34. S. T. Piantadosi, Zipf's word frequency law in natural language: A critical review and future directions. *Psychonomic Bulletin & Review* **21**, 1112–1130 (2014).
35. G. K. Zipf, *Human behavior and the principle of least effort* (Addison-Wesley Press, 1949).
36. A. Tversky, Features of similarity. *Psychological Review* **84**, 327–352 (1977).
37. S. A. Waraich, J. D. Victor, The Geometry of Low- and High-Level Perceptual Spaces. *The Journal of Neuroscience* **44** (2024).
38. M. N. Hebart, C. Y. Zheng, F. Pereira, C. I. Baker, Revealing the multidimensional mental representations of natural objects underlying human similarity judgements. *Nature Human Behaviour* **4**, 1173–1185 (2020).
39. P. Sinha, R. Russell, A Perceptually Based Comparison of Image Similarity Metrics. *Perception* **40**, 1269–1281 (2011).
40. R. Zhang, P. Isola, A. A. Efros, E. Shechtman, O. Wang (2018) The Unreasonable Effectiveness of Deep Features as a Perceptual Metric. in *IEEE/CVF Conference on Computer Vision and Pattern Recognition (CVPR)*, pp 586–595.
41. J. Zujovic, T. N. Pappas, D. L. Neuhoff, Structural Texture Similarity Metrics for Image Analysis and Retrieval. *IEEE T Image Process* **22**, 2545–2558 (2013).

42. S. Fu *et al.*, DreamSim: Learning New Dimensions of Human Visual Similarity using Synthetic Data. *Advances in Neural Information Processing Systems 36 (NeurIPS 2023)* **36** (2023).
43. J. B. Luo, D. Crandall, Color object detection using spatial-color joint probability functions. *IEEE T Image Process* **15**, 1443–1453 (2006).
44. M. J. Swain, D. H. Ballard, Color Indexing. *International Journal of Computer Vision* **7**, 11–32 (1991).
45. J. Domke, Y. Aloimonos (2006) Deformation and Viewpoint Invariant Color Histograms. in *Procedings of the British Machine Vision Conference 2006*, pp 53.51–53.10.
46. T. Valentine, Face-space models of face recognition. *Sci Psych S*, 83–113 (2001).
47. K. Dobs, J. Yuan, J. Martinez, N. Kanwisher, Behavioral signatures of face perception emerge in deep neural networks optimized for face recognition. *Proceedings of the National Academy of Sciences* **120** (2023).
48. M. Q. Hill *et al.*, Deep convolutional neural networks in the face of caricature. *Nature Machine Intelligence* **1**, 522–529 (2019).
49. E. Tousi, M. Mur, The face inversion effect through the lens of deep neural networks. *Proceedings of the Royal Society B: Biological Sciences* **291** (2024).
50. D. Hoiem, A. A. Efros, M. Hebert, Recovering Surface Layout from an Image. *International Journal of Computer Vision* **75**, 151–172 (2007).
51. D. Hoiem, A. A. Efros, M. Hebert, Putting Objects in Perspective. *International Journal of Computer Vision* **80**, 3–15 (2008).
52. V. Hedau, D. Hoiem, D. Forsyth, "Thinking Inside the Box: Using Appearance Models and Context Based on Room Geometry" in Computer Vision – ECCV 2010. (2010), 10.1007/978-3-642-15567-3_17 chap. 17, pp. 224–237.
53. J. Ao, Q. Ke, K. A. Ehinger, Image amodal completion: A survey. *Computer Vision and Image Understanding* **229** (2023).
54. A. Kar, S. Tulsiani, J. Carreira, J. Malik (2015) Amodal Completion and Size Constancy in Natural Scenes. in *IEEE/CVF International Conference on Computer Vision (ICCV)*, pp 127–135.
55. Z. Sadeghi, J. L. McClelland, P. Hoffman, You shall know an object by the company it keeps: An investigation of semantic representations derived from object co-occurrence in visual scenes. *Neuropsychologia* **76**, 52–61 (2015).
56. A. Doerig *et al.*, High-level visual representations in the human brain are aligned with large language models. *Nature Machine Intelligence* **7**, 1220–1234 (2025).
57. A. G. de Varda, M. Petilli, M. Marelli, A distributional model of concepts grounded in the spatial organization of objects. *Journal of Memory and Language* **142** (2025).
58. A. Oliva, A. Torralba, The role of context in object recognition. *Trends in Cognitive Sciences* **11**, 520–527 (2007).
59. O. Sporns, R. F. Betzel, Modular Brain Networks. *Annual Review of Psychology* **67**, 613–640 (2016).
60. S. Fortunato, Community detection in graphs. *Physics Reports* **486**, 75–174 (2010).
61. M. Á. Serrano, M. Boguñá, A. Vespignani, Extracting the multiscale backbone of complex weighted networks. *Proceedings of the National Academy of Sciences* **106**, 6483–6488 (2009).
62. R. Mastrandrea, A. Yassin, H. Cherifi, H. Seba, O. Togni, Backbone extraction through statistical edge filtering: A comparative study. *Plos One* **20** (2025).
63. C. R. Buchanan *et al.*, The effect of network thresholding and weighting on structural brain networks in the UK Biobank. *NeuroImage* **211** (2020).

64. S. E. Favila, H. Lee, B. A. Kuhl, Transforming the Concept of Memory Reactivation. *Trends Neurosci* **43**, 939–950 (2020).
65. M. Penrose, *Random Geometric Graphs* (Oxford University Press, ed. 1st, 2003).
66. R. Diestel, *Graph Theory*, Graduate Texts in Mathematics (Springer Berlin Heidelberg, ed. 6, 2025), 10.1007/978-3-662-70107-2.
67. L. W. Beineke, O. R. Oellermann, R. E. Pippert, The average connectivity of a graph. *Discrete Math* **252**, 31–45 (2002).
68. J. M. Mandler, P. J. Bauer, The cradle of categorization: Is the basic level basic? *Cognitive Dev* **3**, 247–264 (1988).
69. U. Hasson, S. A. Nastase, A. Goldstein, Direct Fit to Nature: An Evolutionary Perspective on Biological and Artificial Neural Networks. *Neuron* **105**, 416–434 (2020).
70. A. X. Chang *et al.*, ShapeNet: An Information-Rich 3D Model Repository. arXiv [Preprint] (2015). http://dx.doi.org/10.48550/arXiv.1512.03012 (accessed June 2024).
71. R. Shinoda, K. Shiohara, "PetFace: A Large-Scale Dataset and Benchmark for Animal Identification" in Computer Vision – ECCV 2024. (2025), 10.1007/978-3-031-72649-1_2 chap. Chapter 2, pp. 19–36.
72. K. M. He, X. Y. Zhang, S. Q. Ren, J. Sun (2016) Deep Residual Learning for Image Recognition. in *IEEE/CVF Conference on Computer Vision and Pattern Recognition (CVPR)*, pp 770–778.
73. K. Wu *et al.*, "TinyViT: Fast Pretraining Distillation for Small Vision Transformers" in Computer Vision – ECCV 2022. (2022), 10.1007/978-3-031-19803-8_5 chap. Chapter 5, pp. 68–85.
74. W. K. Vong, W. Wang, A. E. Orhan, B. M. Lake, Grounded language acquisition through the eyes and ears of a single child. *Science* **383**, 504–511 (2024).
75. C. Suarez-Rivera, E. Linn, C. S. Tamis-LeMonda, From Play to Language: Infants' Actions on Objects Cascade to Word Learning. *Language Learning* **72**, 1092–1127 (2022).
76. J. J. DiCarlo, D. D. Cox, Untangling invariant object recognition. *Trends in Cognitive Sciences* **11**, 333–341 (2007).
77. D. Marr, H. K. Nishihara, Representation and recognition of the spatial organization of three-dimensional shapes. *Proceedings of the Royal Society of London. Series B. Biological Sciences* **200** (1978).
78. T. Poggio, S. Edelman, A network that learns to recognize three-dimensional objects. *Nature* **343**, 263–266 (1990).
79. J. J. DiCarlo, D. Zoccolan, N. C. Rust, How does the brain solve visual object recognition? *Neuron* **73**, 415–434 (2012).
80. F. Jäkel, B. Schölkopf, F. A. Wichmann, Generalization and similarity in exemplar models of categorization: Insights from machine learning. *Psychonomic Bulletin & Review* **15**, 256–271 (2013).
81. R. L. Goldstone, Influences of categorization on perceptual discrimination. *Journal of Experimental Psychology: General* **123**, 178–200 (1994).
82. J. R. Folstein, T. J. Palmeri, I. Gauthier, Category Learning Increases Discriminability of Relevant Object Dimensions in Visual Cortex. *Cerebral Cortex* **23**, 814–823 (2013).
83. C. A. Sanders, R. M. Nosofsky, Training Deep Networks to Construct a Psychological Feature Space for a Natural-Object Category Domain. *Computational Brain & Behavior* **3**, 229–251 (2020).
84. S. Duvdevani-Bar, S. Edelman, Visual Recognition and Categorization on the Basis of Similarities to Multiple Class Prototypes. *International Journal of Computer Vision* **33**, 201–228 (1999).

85. W. W. Sun *et al.*, Canonical Capsules: Self-Supervised Capsules in Canonical Pose. *Advances in Neural Information Processing Systems 34 (NeurIPS)* **34** (2021).
86. J. Wang, Y. Chen, S. X. Yu, "Pose-Aware Self-supervised Learning with Viewpoint Trajectory Regularization" in Computer Vision – ECCV 2024. (2025), 10.1007/978-3-031-72664-4_2 chap. Chapter 2, pp. 19–37.
87. P. K. Kuhl *et al.*, Phonetic learning as a pathway to language: new data and native language magnet theory expanded (NLM-e). *Philosophical Transactions of the Royal Society B: Biological Sciences* **363**, 979–1000 (2007).
88. F. H. Guenther, M. N. Gjaja, The perceptual magnet effect as an emergent property of neural map formation. *The Journal of the Acoustical Society of America* **100**, 1111–1121 (1996).
89. P. Iverson, P. K. Kuhl, Mapping the perceptual magnet effect for speech using signal detection theory and multidimensional scaling. *The Journal of the Acoustical Society of America* **97**, 553–562 (1995).
90. A. Baranes, P.-Y. Oudeyer, Active learning of inverse models with intrinsically motivated goal exploration in robots. *Robotics and Autonomous Systems* **61**, 49–73 (2013).
91. Y. Bengio, J. Louradour, R. Collobert, J. Weston (2009) Curriculum learning. in *Proceedings of the 26th Annual International Conference on Machine Learning*, pp 41–48.
92. M. Xu, S. Yoon, A. Fuentes, D. S. Park, A Comprehensive Survey of Image Augmentation Techniques for Deep Learning. *Pattern Recognition* **137** (2023).
93. T. Chen, S. Kornblith, M. Norouzi, G. Hinton, A Simple Framework for Contrastive Learning of Visual Representations. *Pr Mach Learn Res* **119** (2020).
94. R. Balestriero, L. Bottou, Y. LeCun, The Effects of Regularization and Data Augmentation are Class Dependent. *Adv Neur In* **35** (2022).
95. T. Devries, G. W. Taylor, Improved Regularization of Convolutional Neural Networks with Cutout. ArXiv [Preprint] (2017). http://dx.doi.org/10.48550/arXiv.1708.04552 (accessed September 2025).
96. C. F. G. Dos Santos, J. P. Papa, Avoiding Overfitting: A Survey on Regularization Methods for Convolutional Neural Networks. *ACM Computing Surveys* https://doi.org/10.1145/3510413, Article 123 (2022).
97. G. Zhang, M. Cisse, Y. Dauphin, D. Lopez-Paz (2018) mixup: Beyond Empirical Risk Minimization. in *International Conference on Learning Representations (ICLR)*.
98. C. B. Hilton *et al.*, Acoustic regularities in infant-directed speech and song across cultures. *Nature Human Behaviour* **6**, 1545–1556 (2022).
99. R. Li *et al.*, Why Do Cultures Affect Facial Emotion Perception? A Systematic Review. *Journal of Cross-Cultural Psychology* **56**, 680–710 (2025).
100. L. B. Karasik, K. E. Adolph, S. N. Fernandes, S. R. Robinson, C. S. Tamis-LeMonda, Gahvora cradling in Tajikistan: Cultural practices and associations with motor development. *Child Development* **94**, 1049–1067 (2023).
101. Y. Zhu, B. Zhang, Q. A. Wang, W. Li, X. Cai, The principle of least effort and Zipf distribution. *Journal of Physics: Conference Series* **1113** (2018).
102. G. J. Ross, T. Jones, Understanding the heavy-tailed dynamics in human behavior. *Physical Review E* **91** (2015).
103. A.-L. Barabási, The origin of bursts and heavy tails in human dynamics. *Nature* **435**, 207–211 (2005).
104. B. M. Lake, R. Salakhutdinov, J. B. Tenenbaum, Human-level concept learning through probabilistic program induction. *Science* **350**, 1332–1338 (2015).

105. F. H. Sinz, X. Pitkow, J. Reimer, M. Bethge, A. S. Tolias, Engineering a Less Artificial Intelligence. *Neuron* **103**, 967–979 (2019).
106. C. M. Fausey, S. Jayaraman, L. B. Smith, From faces to hands: Changing visual input in the first two years. *Cognition* **152**, 101–107 (2016).
107. D. M. Regal, D. H. Ashmead, P. Salapatek, The coordination of eye and head movements during early infancy: A selective review. *Behavioural Brain Research* **10**, 125–132 (1983).
108. H. Bloch, I. Carchon, On the onset of eye-head coordination in infants. *Behavioural Brain Research* **49**, 85–90 (1992).
109. H. Yoshida, L. B. Smith, What's in View for Toddlers? Using a Head Camera to Study Visual Experience. *Infancy* **13**, 229–248 (2008).
110. C. Schmitow, G. Stenberg, A. Billard, C. v. Hofsten, Using a head-mounted camera to infer attention direction. *International Journal of Behavioral Development* **37**, 468–474 (2013).
111. S. Bambach, L. B. Smith, D. J. Crandall, C. Yu (2016) Objects in the center: How the infant's body constrains infant scenes. in *2016 Joint IEEE International Conference on Development and Learning and Epigenetic Robotics (ICDL-EpiRob)*, pp 132–137.
112. T. R. Candy *et al.*, Infants' use of eye movements to explore their natural environment. *Journal of Vision* **24** (2024).
113. A. Paszke *et al.*, PyTorch: An Imperative Style, High-Performance Deep Learning Library. *2019 Advances in Neural Information Processing Systems 32 (NeurIPS)* **32** (2019).
114. R. Wightman (2019) PyTorch Image Models. (GitHub).

## Supplementary Information for

A solution to generalized learning from small training sets found in infants' repeated visual experiences of individual objects

**This file includes:**

Extended Methods
Figures S1 to S8 and Supporting Text
Tables S1 to S4 and Supporting Text
SI References

## Extended Methods: Infant Visual Experiences of 8 Object Categories

### Data collection

The corpus of egocentric images during mealtime events is a portion of a larger data set of head-camera videos collected by infants in their homes, $n$ = 100, from 3 weeks to 24 months old (1). The head camera was the commercially available Looxcie 2. The camera has three critical properties relevant to this study: a very lightweight 22g body, built-in recording capacity, and a rechargeable non-heating battery. The camera measures 0.91" x 0.67" x 3.33", has an f2.8 lens with a 75° diagonal field of view (41° vertical FOV, 69° horizontal FOV), and a 2" to infinity depth of focus. The camera recorded at 30 Hz. The camera was attached to a headband that held the camera low on the toddlers' forehead directly above the nose. Infants and toddlers tend to orient eyes and heads in the same direction (bringing them into alignment within 500 msec) when shifting attention (2-5). Studies using head-mounted eye-tracking demonstrate that 80% of gaze by infants and children falls within 20 degrees of the center of the head-camera image (6, 7). Parents were taught to place and operate the head cameras and were asked to collect 6 to 8 hours of daily-life experiences; no experimenters were present in the home during data collection. Parents were told that the study was about early visual experiences and that all visual experiences were relevant. The subcorpus of 87 mealtimes (mean duration of a mealtime = 11.22 min., SD = 11.87) was extracted. Analyses of the objects in view and the object names heard using the mealtime corpus were reported in two prior studies (8, 9).

### Identification of individual instances of the 8 categories

Adults (through Amazon Mechanical Turk) were paid to label the objects in each image (see (9)). From this coding we selected the 8 target objects that had high frequency in the mealtime corpus and whose names were normatively among the first 100 nouns learned by children learning American English (10). The number of instances in an image and the identity of the individual instances were coded by trained adults in the laboratory. The total number $n$ of object instances per category visible in each image was recorded. Unique alphabetical IDs were assigned to each instance across the entire corpora of data (e.g., spoon A, spoon B, spoon C, etc.) for each category and independently for each infant for images with $n < 4$ instances of the target category. For example, consider an image with two spoons in view. Each of these spoons has a fixed identity across the entire data set. Thus, in such an image, coders may identify that spoons B and C are in view, but not A. In this case, $n = 2$ for the category spoon for that particular image. This is performed for each of the 8 target categories and for each image. Images with greater than 4 instances for the target category (e.g., more than 4 spoons within a single image) were not included in the determination of pairwise similarities for that category (e.g., in Fig. 2A); such images are rare and tend to have objects that are too small to identify with certainty. Two independent coders coded 20% of the images; image-to-image agreement on the number of instances in an image and the identity of those instances present both exceeded 84%. Lastly, the unit of analysis is an object occurrence, meaning that an image containing more than one target category is counted as a datapoint for each of those categories.

### Computations for all Figures in the Main Text

**Number of instances per image.** For each category, we count the number of instances, $n$, present in each image. We keep images with at least one instance ($n \geq 1$), and group images with $n \geq 4$ together as they are rare. Next, we obtain proportions by dividing the number of images with $n \in$ $\{1, 2, 3, 4\}$ instances by the total number of images within that category. The resulting proportions per category are the colored dots displayed in Figure S1. To generate the bars, we average the proportions across categories.

**Rank frequency within a mealtime.** For each category, mealtime, and subject, we count the number of occurrences of each unique object ID present during the mealtime. Then, we rank the object IDs by frequency in descending order, where the rank 1 instance is the one present in the largest number of images within a mealtime. Next, we aggregate the total number of occurrences for each rank across mealtimes and subjects and normalize by the sum of counts to obtain the proportions for each rank. The resulting proportions per category are the colored dots displayed in Figure S1. To generate the bars, we average the proportions across categories.

**Rank frequency across mealtimes (Fig. 1, Panel C).** Occurrences and ranks are now computed across the entire corpora of data for a given infant (i.e., across mealtime events) rather than within mealtime events. Thus, for a given category, the rank 1 instance is the most frequent object in that category across all recorded experiences of a given infant, and similarly for the remaining ranks. To test for significance against the null distribution, however, we aggregate the raw counts across categories and then normalize by the sum of counts to obtain the exact proportions (i.e., a distribution that adds up to 1.0), rather than averaging the proportions across categories. To generate the null distribution, we uniformly sample an equal total number of counts for each subject and category from the available object IDs for each subject. This is performed 500 times with different seeds for the random number generator (0 through 499) and the raw counts are aggregated across categories and normalized by rank to generate the null frequency distribution (displayed as a smoothed red line in the figure). To test for significance, we use a single seed to have an equal number of counts as in the observed infant data and perform a KS test on the observed (discrete) counts. The number of counts (data points) used for the KS test is 20,339 for each of the observed infant and null distributions, for a total of 40,678 samples.

**Similarity analyses.** Raw RGB histogram similarities were computed using the open-source OpenCV library (11). Histogram correlation $c_{i,j}$ between each pair of images $(i, j)$ is computed in RGB space using 8 histogram bins per channel, for a total of 512 bins. The resulting value $-1 \leq c_{i,j} \leq 1$ is a similarity score, with a higher value denoting a higher similarity. In our results, $c_{i,j} \in (-0.04, 1)$ and thus we set our figure axes to the range $[0, 1]$ for visualization purposes and use all data for the measures. GIST (12), CLIP (13), and Perception-Encoder (14) image feature similarities were measured using cosine similarity. GIST features employed a Gabor filter with 3 scales with 8, 8, and 4 orientations respectively. We used the ViT-Base version of CLIP, and the PE-Spatial-L (a ViT-Large) version of Perception-Encoder. PE-Spatial captures both spatial correspondences useful for segmentation and category-level information, and we compute cosine similarity on its CLS token. For each category and comparison type, infant subjects with fewer than 10 total images were excluded. We provide a deeper discussion on different similarity measures in the SI.

**Infant and null similarity density (Fig. 2, Panel A, *Left*).** We illustrate aggregated pairwise similarities across all categories, but the pairwise similarity comparisons are still performed only within-category. We remove duplicate pairwise datapoints that arise after merging categories. The box plots illustrate the 0.25, 0.5 (median), and 0.75 quartiles. There are 1,188,150 datapoints in total for the infant distribution. To generate the null distribution, we selected all instances that had at least 5 images, for each category and subject. Then we sampled an equal number of images per instance based on the instance with the fewest images. We filtered out subjects with fewer than 10 resulting images across all instances. Next, for each category, we identified the subject who had the fewest images and sampled uniformly (across instances) that number of images from each subject. Lastly, we compute pairwise similarities among the resulting set of images. This yielded a null distribution with 97,118 datapoints, for a total of 1,285,268. We computed their entropy by discretizing the range $[0, 1.04]$ into 100 bins. The bars in Fig. S3 illustrate the entropy of all pairwise similarities (1,285,268 datapoints), and the colored dots illustrate the entropy of each category independently. In the latter case, the Infant and Null distribution contain 1,519,564 and 97,417 datapoints instead of 1,188,150 and 97,118 datapoints, respectively, as duplicates across categories were not removed.

**Infant similarity density by group (Fig. 2, Panel A, *Right*).** We used an analogous procedure to Fig. 2A (*Left*), but now grouped by rank. For each group, we filtered out subjects with less than 10 images and removed duplicate pairwise datapoints that arose after merging categories. The total number of datapoints across all categories and groups was now 1,037,730. The similarity density by infant is illustrated in Fig. S3. The similarity density by category is also illustrated in Fig. S3, which contains a total of 1,117,337 datapoints as the categories were treated independently and thus duplicate pairwise datapoints were not removed.

**Graph networks (Fig. 2, Panels B, C).** Graphs were created for each infant and category, where each image served as a node connected by edges based on similarity. A linear threshold is applied to the edges for analysis (15-17), and we report here those that represent the top $\tau = 10\%$ of measured similarities. We observe consistent results with higher and lower $\tau$ values (Fig. S2CD). The three null hypotheses were instantiated as follows.

(i) The Shuffled Labels null shuffles the subject-category image labels. This is a bijective mapping with a single constraint: no image label gets mapped to itself. For example, image 0 from subject "A" and category "Bottle" may be mapped to subject "B" and category "Cup," essentially being assigned to a different network. This preserves the underlying spatial distribution of RGB pixels (the original images) and their frequency distribution (due to the bijection) but breaks temporal autocorrelation as nearby-in-time frames from a single source network can be mapped to separate target networks. Pairwise similarities are re-computed within each target network. Note that we shuffled subject-category labels, as opposed to only category labels, because (a) Both category membership and their broader visual context likely play a role at this early stage of category learning, justifying the inclusion of subject labels as part of the null hypotheses; and (b) More than half of the images contain more than one target category in view (e.g., a cup, a table, and a window; Fig. S1A), and thus solely shuffling their category labels while fixing the subject labels (and their visual context) would be less meaningful.

(ii) The Blocks of Shuffled Labels null also shuffles subject-category image labels as in the Shuffled Labels null, but with the added constraint that it is only allowed to shuffle contiguous blocks of $T$ images corresponding to the same source mealtime. That is, while blocks can be shuffled across mealtimes and subject-category networks, the blocks themselves are generated within each mealtime (or otherwise they would not be truly contiguous in time). We retain remainder blocks of size $K \leq T$. For example, when $T = 6$, a mealtime with 64 images will contribute ten blocks of size $T = 6$ and one block of size $K = 4$. The shuffling bijection ensures that source-target blocks are of the same size; e.g., the remainder block of size $K$ will be assigned to another block of size $K$. This requires at least two different blocks for any given size (a source and a target), which holds empirically for all values of $T$ used in our work. With our default value of $T = 6$, the proportion of blocks of size exactly $T$ is 0.938. We report more detailed block-level statistics in the SI. Since the original images were sampled at 0.2Hz, $T = 6$ corresponds to approximately 30 seconds. Note, however, that some images could be skipped during the annotation process if they contained none of the eight target categories. This temporal duration approximation suffices for our purposes, and we provide smaller and larger $T$ values in the SI which show consistent and monotonic results. Lastly, note that the Shuffled Labels null in (ii) is a special case of the Blocks of Shuffled Labels null in which $T = 1$ (each image label is shuffled independently).

(iii) The Geometric null places nodes uniformly at random within the unit square, which is a bounded space like our similarity metric, and connects nodes that fall within a given radius of each other (a similarity threshold given by the corresponding infant graph). As the name indicates, this null hypothesis is based solely on geometry and does not involve images (18).

In all three instantiations, each null network has the same number of nodes as its corresponding infant network.

Regarding metrics, the proportion of isolated nodes was computed for each infant-by-category network and then averaged across networks. For average node degree, isolated nodes contribute a degree of zero. Average node connectivity was computed using the average_node_connectivity function in the open-source NetworkX library (19). Metrics that require connected networks, such as average connectivity, were computed for each connected component and then a weighted average across components was calculated, where the weight is the number of node pairs in each component. Single-node components (i.e., isolated nodes) were not included in such network metrics. The null networks were generated for each subject and category, excluding those with fewer than 10 nodes (images). For each of the three Nulls and observed Infant data, there were 94 networks with at least 10 nodes (376 networks in total). For each of the R1-R1 and O-O subgroups, there were 60 such networks, and for the R1-O subgroup there were 48 networks. The linear mixed-effects models had distribution (Infant or Nulls) and categories as fixed effects and optionally an intercept for the infant subjects. For illustration purposes, graph measures such as average degree and average node connectivity were averaged first across infants for each category, then normalized to the range $[0.0, 1.0]$ by max-division between infant and null graph pairs, and lastly averaged across categories. The error bars illustrate corpus-level standard deviation. The un-normalized values are illustrated in the SI (Fig. S2).

**Statistical analysis of aggregate measures.** We used the large number of continuous data points specified above for several statistical measures, such as to compare means between two distributions using Welch's t-test. However, other measures such as entropy represent a single aggregate number for several thousand or million datapoints. As such, it is not possible to compute test statistics on a single observation, so we used permutation tests. To do so, we first computed the measure (e.g., entropy) for the entire corpus of data and recorded the difference between Infant and Null. Under the null hypothesis, we should observe similar differences simply by chance, regardless of the data distribution label (Infant or Null). Hence, for $n$ iterations, we sampled subsets of size $K$, randomly permuted the labels of the $K$ sampled datapoints, and computed the difference in measures (e.g., entropy) for the permuted subsets. The resulting p-value is the fraction of times that such an extreme difference is observed by chance. For the entropy permutation tests, we used $n =$ 9,999 permutations and sampled 25% of the datapoints at each iteration. For the computational data that follows, we used $n$ = 999 and sampled 20% of the datapoints, as the corpus is larger. Lastly, as complementary non-permutation tests, we also computed the entropy of the $n$ unpermuted subsamples and performed a Welch's two-sample t-test and a Wilcoxon rank-sum test on these $n$ estimates. These yielded results consistent with the permutation test for both the observed infant data and computational data ($p < 0.001$).

**Supplementary Information Text and Additional Figures and Tables**

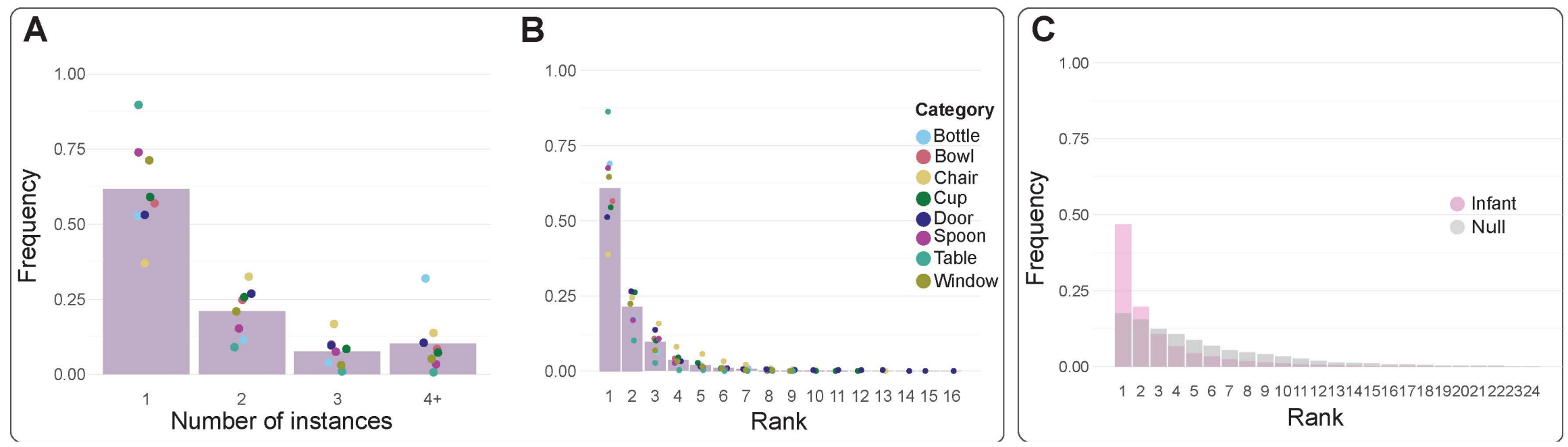

**Fig. S1.** Several characterizations of infant mealtime experiences. **(A)** Number of instances of a given target category per image for each category in view. Infant experiences tend to contain one category member at a time within an image, and thus within a single moment of time (e.g., one cup, one bowl). However, individual images can contain more than one category at a time (e.g., a cup and a bowl). The proportion of images with more than one category in view is 0.53. **(B)** Frequency distribution by rank within a mealtime; i.e., the ranks are determined within each mealtime and then aggregated across mealtimes. We see that the Rank 1 instance within a mealtime tends to dominate. The bars illustrate the average across categories. **(C)** Frequency distribution by rank across mealtimes. Here, we show the non-smoothed null distribution instead of a smoothed version as in Fig. 1.

## Fig. S2

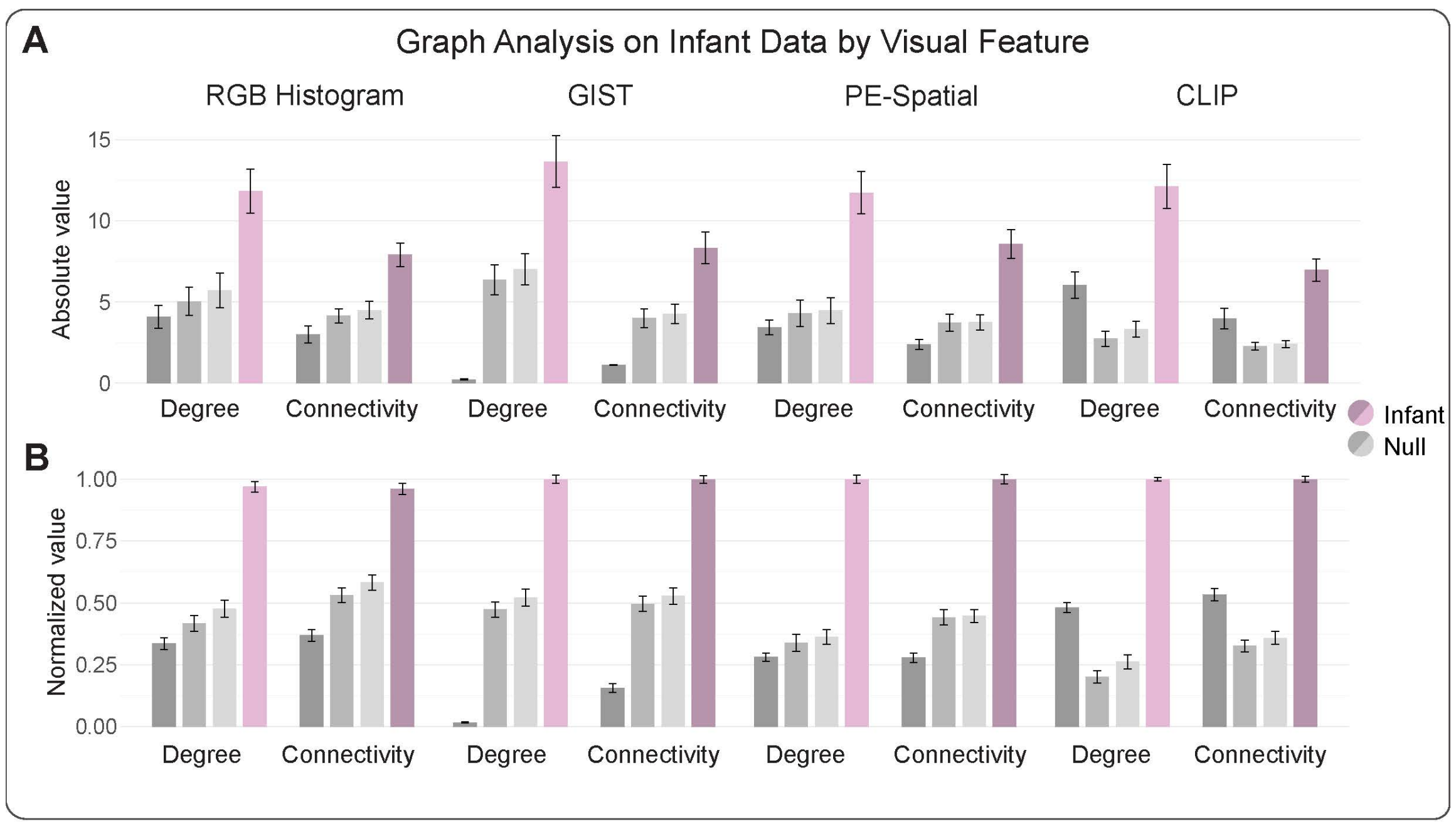

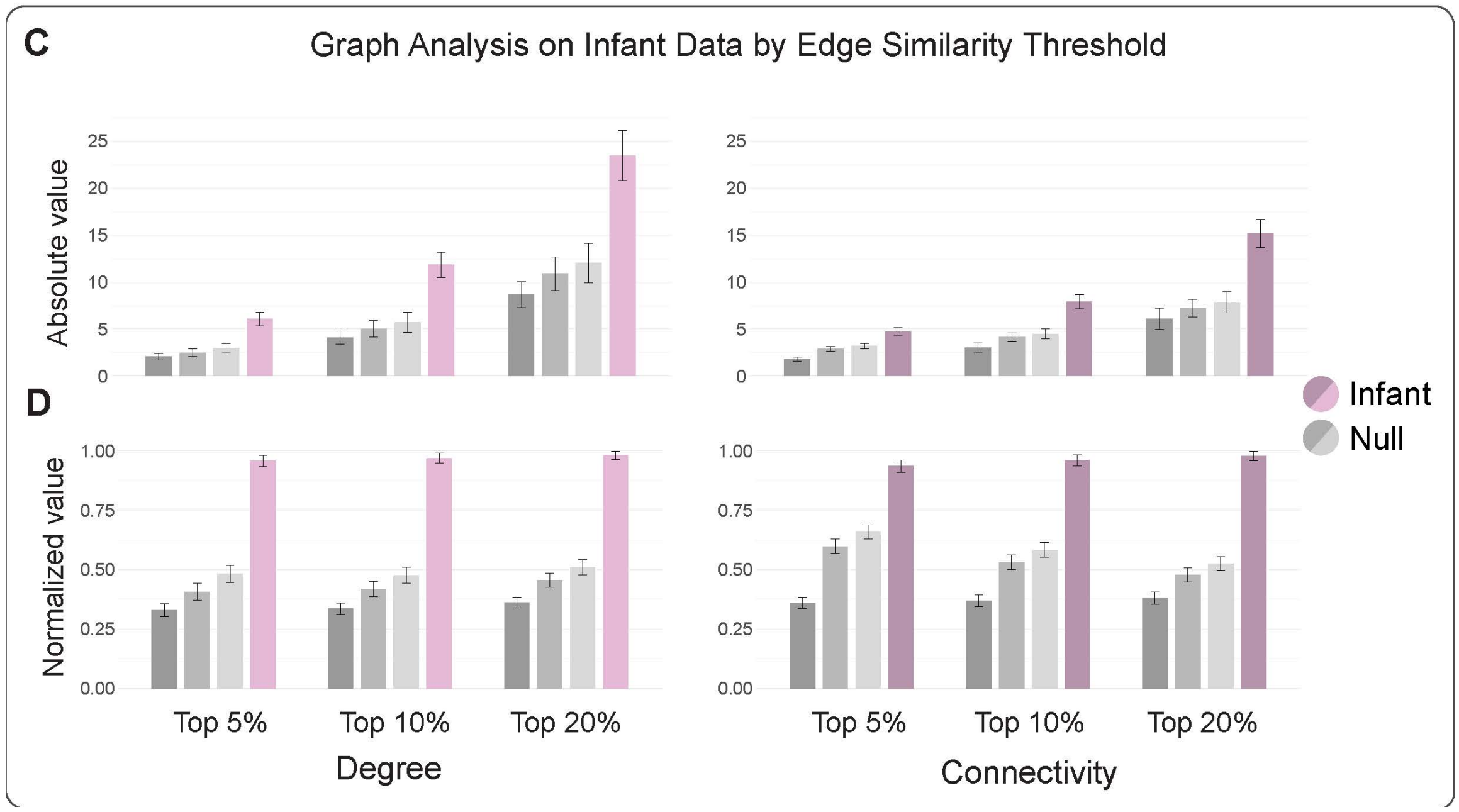

**Fig. S2.** Graph-theoretic measures of Infant and Null networks under different visual features and edge similarity thresholds. **(A)** Graph-theoretic measures using RGB histograms (raw low-level statistics), GIST (spatial layout and edge statistics; Oliva and Torralba (12)), PE-Spatial (spatial structure and semantic categories; Bolya*, et al.* (14)), and CLIP (purely semantic; Radford*, et al.* (13)).

Specifically, GIST features were designed to capture whole scene spatial layout in terms of coarse statistics without encoding category-relevant information, and thus are intermediate in abstraction between raw RGB histograms and category-level representations. CLIP features, derived from a model trained to align images with natural language descriptions, reflect higher-level semantic content and are sensitive to category membership. PE-Spatial features also capture semantic structure but additionally encode spatial correspondences that are useful for segmentation. The appropriate level of abstraction for quantifying image similarity depends on the type of stimuli – textures, complex scenes, faces, objects on clean backgrounds, substances, letters. This is well recognized in both human (20-22) and computer vision research (23-26); in the latter, additional factors such as model architecture, pre-training data, and training objective further determine which image properties a given model captures. Because infant egocentric images contain the content, context, and format of typical photographs for which these measures were derived, all measures show consistent patterns.

For the GIST, CLIP, and PE-Spatial features, we use cosine similarity as our similarity metric between pairs of images (SI Methods). For CLIP and PE-Spatial, cosine similarity is the standard choice (13, 14). For GIST features, either cosine similarity or Euclidean distance is applicable. The distribution of GIST cosine similarities is strongly skewed toward 1 (maximum similarity), which accounts for the larger gap between observed Infant and Geometric Null network connectivity measures visible. Euclidean distance yields a more spread-out distribution while preserving the same qualitative patterns; however, because Euclidean distance is unbounded, it is incompatible with the Geometric Null network comparison, in which nodes are placed uniformly at random within the unit square. We therefore retain cosine similarity (a bounded metric) for GIST features, as well as for CLIP and PE-Spatial (SI Methods).

The three grey (leftmost) bars within a group represent the Geometric, Shuffled Labels, and Blocks of Shuffled Labels nulls, respectively. The number of datapoints per bar and per visual feature (individual infant-by-category networks) is 94, 75, 97, and 93, respectively for each visual feature. For subjects with a low number of nodes in a category, the null networks can be disconnected based on the similarity threshold from the corresponding observed infant network. In such a case, we remove the infant-null network pair for such a subject and category, and thus the number of datapoints can vary depending on the visual feature used. We use a darker color for the "connectivity" infant bars for ease of visualization. The error bars illustrate standard error.

**(B)** Displays relative rather than absolute values. **(C)** Graph measures using RGB histogram features under three different thresholds for drawing an edge between two images: Top 5, 10, and 20% greatest similarities. The thresholds yield similar patterns. High thresholds for connectivity are standard in graph-theoretic analyses as they have been repeatedly shown to lead to patterns that are reliably predictive of other results, and thus meaningful (15-17, 27). The three grey (leftmost) bars within a group represent the Geometric, Shuffled Labels, and Blocks of Shuffled Labels null, respectively. The number of datapoints per bar and per threshold (individual infant-by-category networks) is 87, 94 (as in Panel A,B), and 97, respectively for each threshold. We use a darker color for the "connectivity" infant bars for ease of visualization. The error bars illustrate standard error. **(D)** Displays relative rather than absolute values.

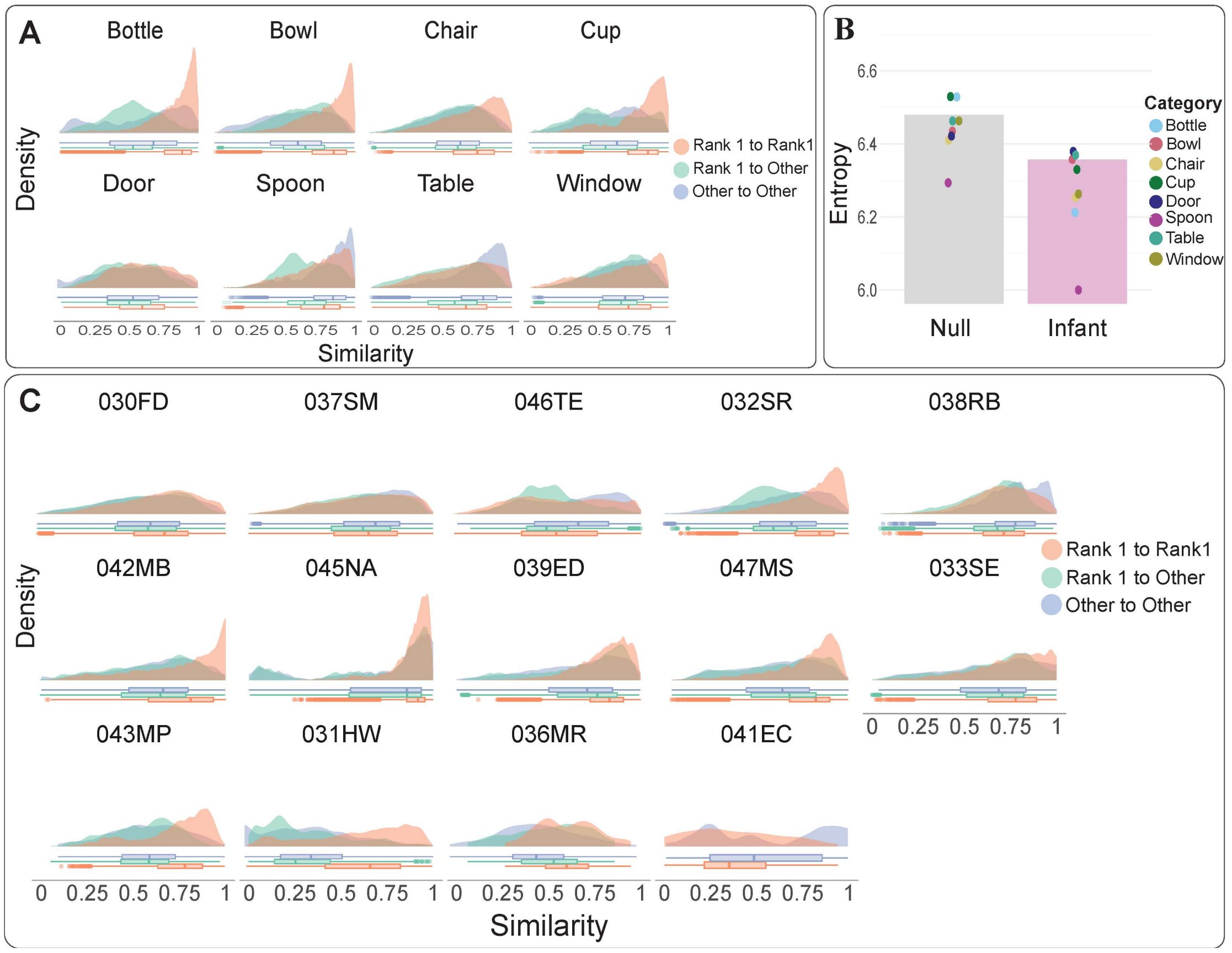

**Fig. S3.** Similarity measures on observed Infant data using RGB histogram features. **(A)** Observed pairwise similarity density by category, partitioned into three groups of pairs: within Rank 1 pairs (R1-R1), Rank 1 to others (R1-O), and all others (non-Rank 1) to each other (O-O). There are 1,117,337 datapoints in total. **(B)** Entropy of pairwise similarities for each distribution, discretized using 100 bins. The bars illustrate the entropy of all pairwise similarities (1,285,268 datapoints), and the colored dots illustrate the entropy of each category independently. The latter contains a total of 1,616,981 datapoints, because categories are treated independently and thus duplicates across categories are not removed for the Infant distribution. **(C)** Observed pairwise similarity density by infant, partitioned into three groups of pairs: within Rank 1 pairs (R1-R1), Rank 1 to others (R1-O), and all others (non-Rank 1) to each other (O-O). There are 1,037,730 datapoints in total, as duplicates across categories were removed as in Fig. 2A (SI Methods). The panels corresponding to each infant are sorted in descending order left to right, top to bottom, based on the amount of data contributed by that infant. That is, the infant on the top left (030FD) contributed the most data in our study, and the infant on the bottom right (041EC) contributed the least. For all similarity analyses, we required at least 10 datapoints to contribute data to an analysis (SI Methods). Subject 041EC does not have R1-O datapoints because none of their categories had both a sufficiently large (at least 10 images per group) Rank1 image group and a sufficiently large Other image group within the same category. Specifically, the category Chair was this infant's only category that contained Other instances, but it contained less than 10 Rank1 Chair images. Thus,

this infant's category Chair only contributed O-O pairs, and therefore this infant only contributed R1-R1 and O-O pairs overall.

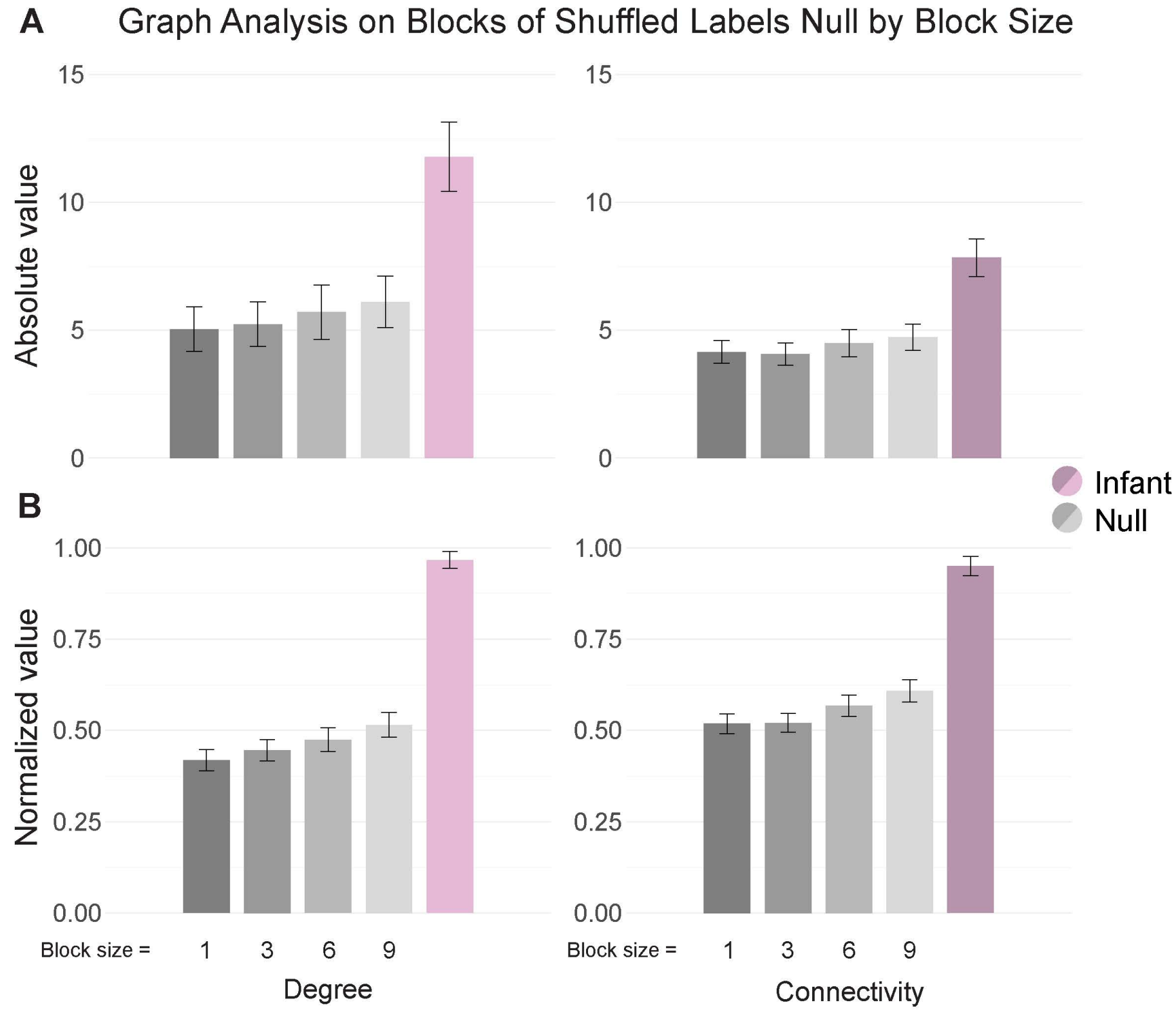

**Fig. S4.** Graph-theoretic measures for the Blocks of Shuffled Labels null graphs under different block sizes using RGB histogram features (SI Methods). **(A)** Displays absolute values. **(B)** Displays relative rather than absolute values. As expected, increasing the block size tends to monotonically increase the measured connectivity values. The leftmost bar, representing a block size of 1, is the special case of the Shuffled Labels null. The number of datapoints per bar (individual infant-by-category networks) is 94. We use a darker color for the "connectivity" infant bars for ease of visualization. The error bars illustrate standard error.

**Fig. S5-S8**

**Graph connectivity measures by infant and visual feature.** Graph measures for the observed infant data and the three nulls – Geometric, Shuffled Labels, and Blocks of Shuffled Labels – for each infant. Each visual feature (e.g., RGB histogram, CLIP) is illustrated in a separate figure. The panels corresponding to each infant are sorted in descending order left to right, top to bottom, based on the amount of data contributed by that infant.

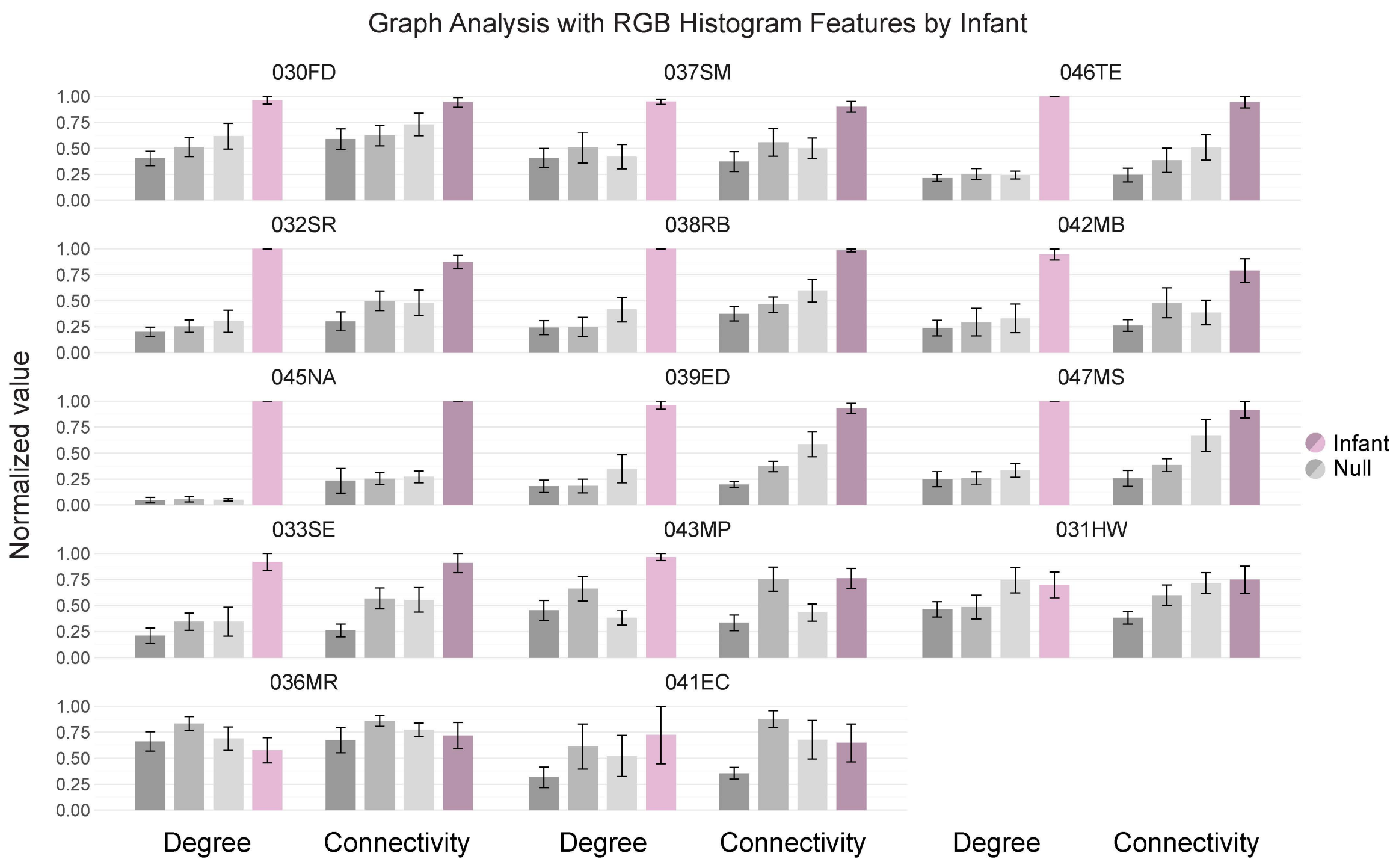

**Fig. S5.** Graph-theoretic measures for each infant using RGB histogram features. The panels corresponding to each infant are sorted in descending order left to right, top to bottom, based on the amount of data contributed by that infant. That is, the infant on the top left (030FD) contributed the most data in our study, and the infant on the bottom right (041EC) contributed the least. The resulting measures yield consistent patterns across infants, and especially for those that have the most data. The three grey (leftmost) bars within a group represent the Geometric, Shuffled Labels, and Blocks of Shuffled Labels null respectively. We use a darker color for the "connectivity" infant bars for ease of visualization. The error bars illustrate standard error.

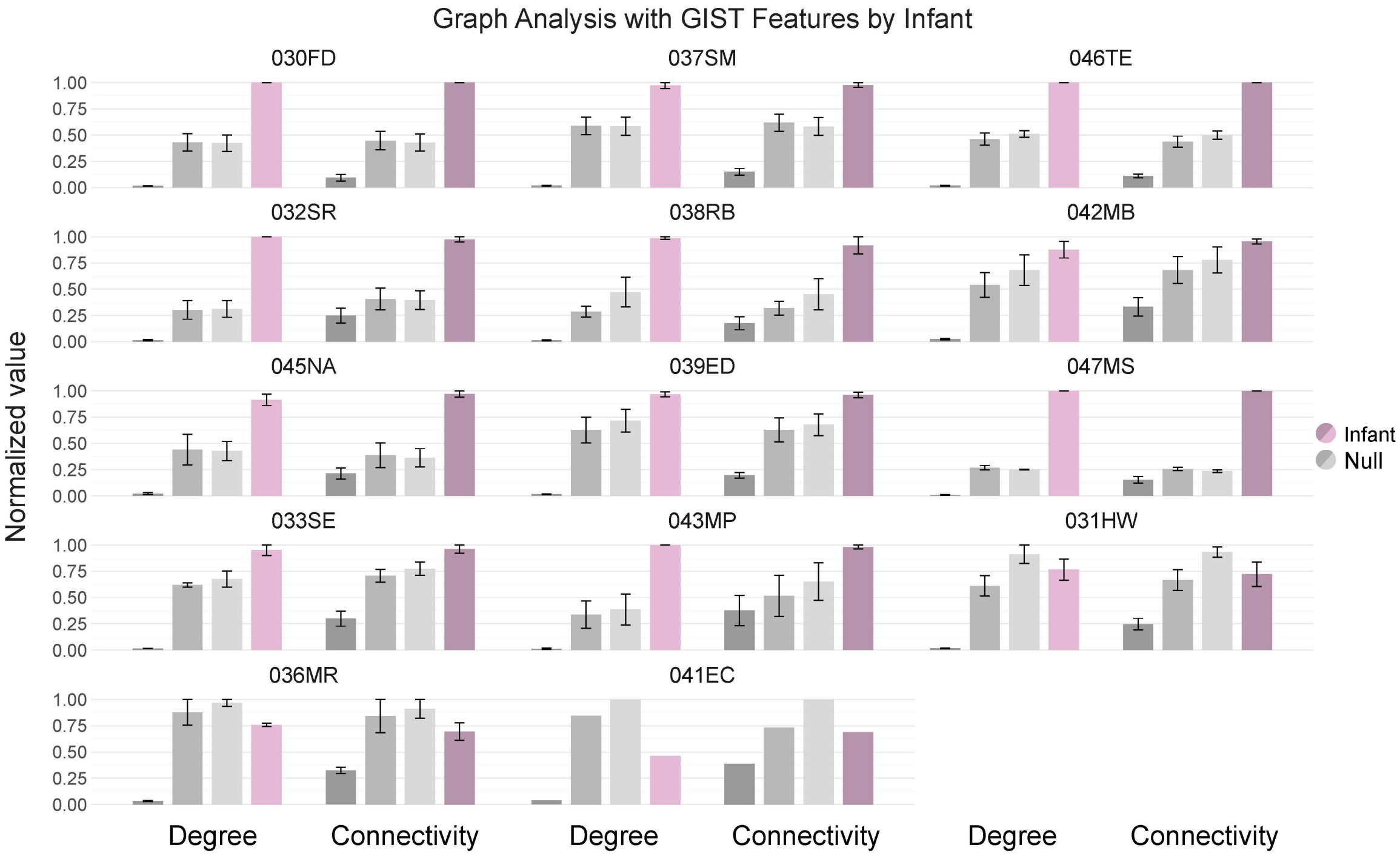

**Fig. S6.** Graph-theoretic measures for each infant using GIST features. The panels corresponding to each infant are sorted in descending order left to right, top to bottom, based on the amount of data contributed by that infant. That is, the infant on the top left (030FD) contributed the most data in our study, and the infant on the bottom right (041EC) contributed the least. Infant 041EC only contributed one graph when using GIST features (see the caption of Fig. S2A) and thus no error bars are shown. The resulting measures yield consistent patterns across infants, and especially for those that have the most data.The three grey (leftmost) bars within a group represent the Geometric, Shuffled Labels, and Blocks of Shuffled Labels null respectively. We use a darker color for the "connectivity" infant bars for ease of visualization. The error bars illustrate standard error.

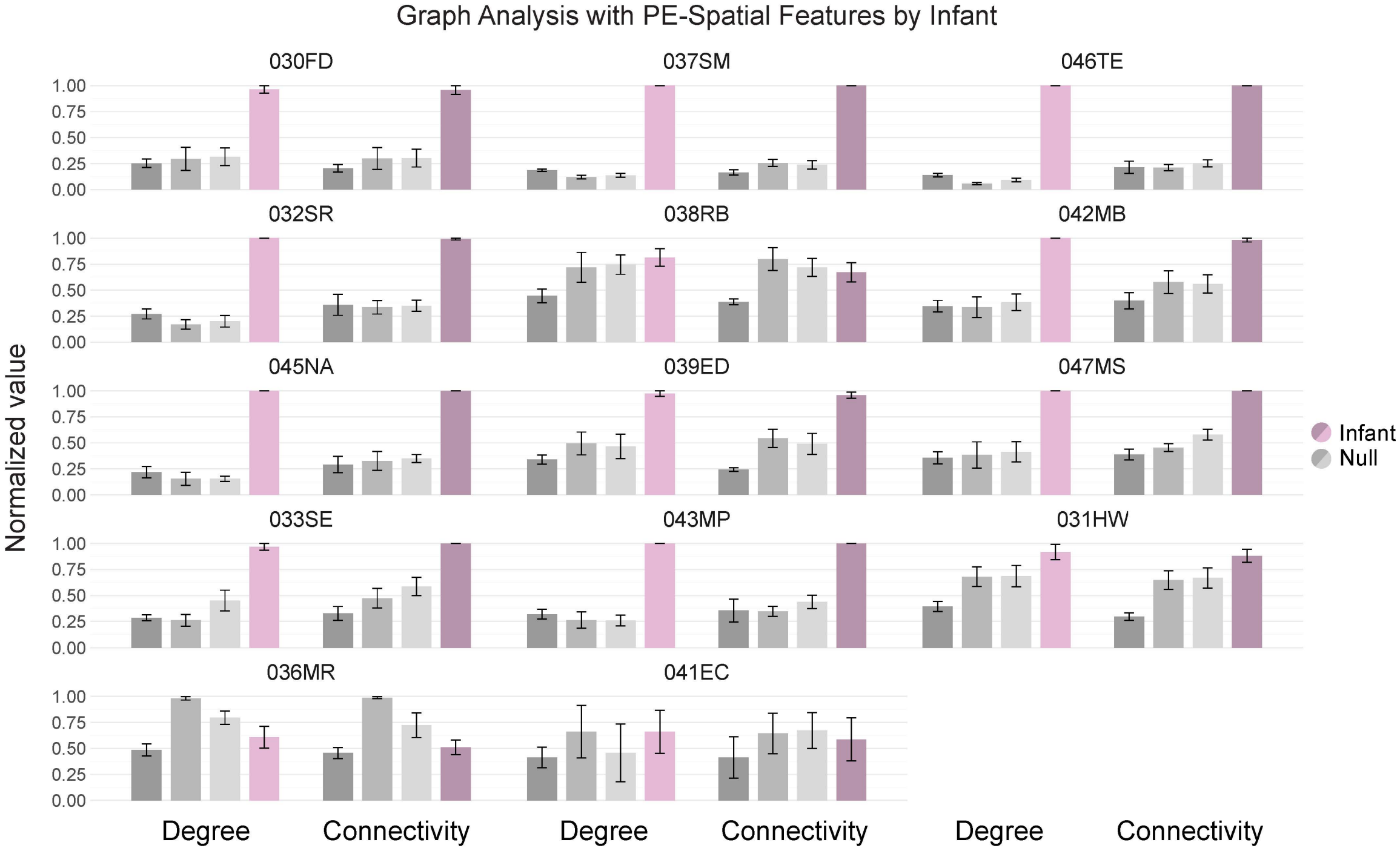

**Fig. S7.** Graph-theoretic measures for each infant using PE-Spatial features. The panels corresponding to each infant are sorted in descending order left to right, top to bottom, based on the amount of data contributed by that infant. That is, the infant on the top left (030FD) contributed the most data in our study, and the infant on the bottom right (041EC) contributed the least. The resulting measures yield consistent patterns across infants, and especially for those that have the most data. The three grey (leftmost) bars within a group represent the Geometric, Shuffled Labels, and Blocks of Shuffled Labels null respectively. We use a darker color for the "connectivity" infant bars for ease of visualization. The error bars illustrate standard error.

**Fig. S8.** Graph-theoretic measures for each infant using CLIP features. The panels corresponding to each infant are sorted in descending order left to right, top to bottom, based on the amount of data contributed by that infant. That is, the infant on the top left (030FD) contributed the most data in our study, and the infant on the bottom right (041EC) contributed the least. The resulting measures yield consistent patterns across infants, and especially for those that have the most data. The three grey (leftmost) bars within a group represent the Geometric, Shuffled Labels, and Blocks of Shuffled Labels null respectively. We use a darker color for the "connectivity" infant bars for ease of visualization. The error bars illustrate standard error.

**Table S1.** Block size statistics for the Blocks of Shuffled Labels null graphs.

| **BLOCK SIZE *T* (IN FRAMES)** | **PROP. OF BLOCKS OF SIZE EXACTLY *T*** |
|---|---|
| 1 (Shuffled Labels null) | 1.0 |
| 3 | 0.975 |
| 6 | 0.938 |
| 9 | 0.899 |

For all the block sizes $T$ used in this work, there are at least 0.899 (out of 1.0) blocks of exactly that size. For $T = 6$, our default value, there are 0.938 blocks of size exactly 6 frames (originally sampled at 0.2Hz, SI Methods). We retain remainder blocks of size less than $T$ (SI Methods), meaning that there are 0.062 of such blocks when $T = 6$. The first row indicates that the Shuffled Labels null is the special case of the Blocks of Shuffled Labels null when $T = 1$, in which each image label is shuffled independently.

**Tables S2 and S3: Similarity structure of the training sets for the machine learning experiments.** We generated training sets in two ways, by oversampling variable views of individual instances, which we expected to create Lumpy training sets, and by uniformly sampling experiences from all available instances, which should generate Uniform training sets. We measured the similarity structure of the training sets under the same connectivity measures and visual features used for the infant data.

The success of machine learning models is in finding structure in the training data. The results of the experiments indicate that oversampling variable images of some instances leads to a similarity structure that is measurably Lumpy by appropriate measures and that is associated with better generalization from very small training sets, and thus a structure findable by these models.

**Table S2.** Graph-theoretic measures for the ShapeNet dataset.

| FEATURE | PROP. ISOLATED | | DEGREE | | CONNECTIVITY | |
|---|---|---|---|---|---|---|
| | Uniform | Lumpy | Uniform | Lumpy | Uniform | Lumpy |
| **RGB HISTOGRAM** | 0.0814 ± .08 | **0.0799 ± .07** | 21.0 ± 4.7 | **30.0 ± 0.7** | 12.0 ± 3.2 | **20.7 ± 8.2** |
| **GIST** | **0.0222 ± .04** | 0.0272 ± .03 | 26.9 ± 3.0 | **28.4 ± 3.2** | 15.8 ± 1.3 | **17.1 ± 3.3** |
| **PE-SPATIAL** | **0.0128 ± .01** | 0.0132 ± .01 | 24.0 ± 3.6 | **30.0 ± 0.7** | 12.1 ± 5.2 | **16.3 ± 3.8** |
| **CLIP** | 0.0237 ± .02 | **0.0197 ± .01** | 27.7 ± 3.2 | **29.2 ± 2.0** | 14.1 ± 1.6 | **14.2 ± 2.8** |

The graph theoretic measures of ShapeNet (28) show the Lumpy structure for the data set with an infant-like oversampling of instances relative to uniform sampling when measured by RGB histograms and PE-Spatial features. The lumpiness of the Lumpy versus Uniform datasets is, at best, weakly evident for GIST and CLIP. This is not surprising because the ShapeNet images show object cutouts with no background and both GIST and CLIP features are whole-scene descriptors, GIST in terms of spatial layout independent of content (12) and CLIP (13) in terms of features predictive of semantic categories (29). Each cell shows the mean (± SD) across 6 categories. The highest value between Uniform and Lumpy pairs is bolded. "Prop. Isolated" stands for Proportion of isolated nodes.

**Table S3.** Graph-theoretic measures for the PetFace dataset.

| FEATURE | PROP. ISOLATED | | DEGREE | | CONNECTIVITY | |
|---|---|---|---|---|---|---|
| | Uniform | Lumpy | Uniform | Lumpy | Uniform | Lumpy |
| **RGB HISTOGRAM** | 0.171 ± .06 | **0.152 ± .03** | 10.8 ± 1.8 | **11.3 ± 1.7** | 6.8 ± 1.0 | **6.9 ± 1.2** |
| **GIST** | 0.258 ± .07 | **0.257 ± .10** | **10.5 ± 3.1** | 9.70 ± 2.3 | **6.7 ± 2.1** | 6.1 ± 1.2 |
| **PE-SPATIAL** | 0.233 ± .04 | **0.219 ± .05** | 11.3 ± 1.6 | **12.0 ± 0.6** | 6.2 ± 0.8 | **7.3 ± 0.8** |
| **CLIP** | 0.224 ± .08 | **0.219 ± .12** | **12.3 ± 0.5** | 9.80 ± 2.8 | **7.8 ± 0.5** | 6.7 ± 1.4 |

PE-Spatial reveals a Lumpy similarity for the animal faces of PetFace (30), but the other features do not. This is expected. The faces of all the animal categories are similar in color (shades of grey, brown and white) and texture, limiting the value of RGB histograms. Furthermore, as has been repeatedly shown for human face recognition, the spatial configurations of faces are distinct from scenes and objects, and thus GIST measures first proposed for scenes do not capture their pattern. Likewise, CLIP features are not designed for faces, human or otherwise. There is a large literature on the specialness of the human face representation supporting these conclusions (31-37). We know of no literature on animal faces. PE-Spatial with its emphasis on spatial correspondences and ability to find segments in the image (the different colored regions of eyes, nose, mouth, etc.) shows the Lumpy structure relative to uniform sampling. Each cell shows the mean (± SD) across 6 categories. The highest value between Uniform and Lumpy pairs is bolded. "Prop. Isolated" stands for Proportion of isolated nodes.

**Table S4.** Robustness test of ResNet-50 generalization accuracy on ShapeNet.

| DATASET | ACCURACY (8 RUNS) | ACCURACY (80 RUNS) |
|---|---|---|
| **UNIFORM** | 0.5727 ± 0.03 | 0.5796 ± 0.03 |

Training deep learning models inherently involves randomness (38-40), especially when training models from scratch and on very small datasets. Variations across different training runs of the same model with the same dataset happen even if the pseudo-random number generator's seeds are fixed, due to the inherent non-determinism of low-level GPU hardware operations (e.g., CUDA atomic operations running in parallel that are scheduled in a different order, (41-43)). To test whether 8 runs are robust enough for the ResNet (44) experiments in the Main Text, we trained an additional 80 runs with a ResNet-50 model on the Uniform training set of ShapeNet. We observe similar mean accuracies (± SD) in both cases; i.e., within ~0.007 accuracy points of each other (out of 1.0). This indicates that the variance seen at 8 runs is not due to a need for additional runs but instead due to the well-known non-determinism of the components mentioned above. Thus, we perform 8 runs for each ResNet trained on ShapeNet in the Main Text.

**SI References**

1. C. M. Fausey, S. Jayaraman, L. B. Smith, From faces to hands: Changing visual input in the first two years. *Cognition* **152**, 101–107 (2016).
2. D. M. Regal, D. H. Ashmead, P. Salapatek, The coordination of eye and head movements during early infancy: A selective review. *Behavioural Brain Research* **10**, 125–132 (1983).
3. H. Bloch, I. Carchon, On the onset of eye-head coordination in infants. *Behavioural Brain Research* **49**, 85–90 (1992).
4. H. Yoshida, L. B. Smith, What's in View for Toddlers? Using a Head Camera to Study Visual Experience. *Infancy* **13**, 229–248 (2008).
5. C. Schmitow, G. Stenberg, A. Billard, C. v. Hofsten, Using a head-mounted camera to infer attention direction. *International Journal of Behavioral Development* **37**, 468–474 (2013).
6. S. Bambach, L. B. Smith, D. J. Crandall, C. Yu (2016) Objects in the center: How the infant's body constrains infant scenes. in *2016 Joint IEEE International Conference on Development and Learning and Epigenetic Robotics (ICDL-EpiRob)*, pp 132–137.
7. T. R. Candy *et al.*, Infants' use of eye movements to explore their natural environment. *Journal of Vision* **24** (2024).
8. E. M. Clerkin, L. B. Smith, Real-world statistics at two timescales and a mechanism for infant learning of object names. *Proceedings of the National Academy of Sciences* **119** (2022).
9. E. M. Clerkin, E. Hart, J. M. Rehg, C. Yu, L. B. Smith, Real-world visual statistics and infants' first-learned object names. *Philosophical Transactions of the Royal Society B: Biological Sciences* **372** (2017).
10. M. C. Frank, M. Braginsky, D. Yurovsky, V. A. Marchman, Wordbank: an open repository for developmental vocabulary data. *Journal of Child Language* **44**, 677–694 (2016).
11. G. Bradski, The OpenCV library. *Dr Dobbs J* **25**, 120–+ (2000).
12. A. Oliva, A. Torralba, Modeling the Shape of the Scene: A Holistic Representation of the Spatial Envelope. *International Journal of Computer Vision* **42**, 145–175 (2001).
13. A. Radford *et al.* (2021) Learning Transferable Visual Models From Natural Language Supervision. in *International Conference on Machine Learning (ICML)* (Proceedings of Machine Learning Research), pp 8748–8763.
14. D. Bolya *et al.* (2025) Perception Encoder: The best visual embeddings are not at the output of the network. in *Advances in Neural Information Processing Systems (NeurIPS)*.
15. M. Á. Serrano, M. Boguñá, A. Vespignani, Extracting the multiscale backbone of complex weighted networks. *Proceedings of the National Academy of Sciences* **106**, 6483–6488 (2009).
16. R. Mastrandrea, A. Yassin, H. Cherifi, H. Seba, O. Togni, Backbone extraction through statistical edge filtering: A comparative study. *Plos One* **20** (2025).
17. C. R. Buchanan *et al.*, The effect of network thresholding and weighting on structural brain networks in the UK Biobank. *NeuroImage* **211** (2020).
18. M. Penrose, *Random Geometric Graphs* (Oxford University Press, ed. 1st, 2003).
19. A. A. Hagberg , D. A. Schult , P. J. Swart (2008) Exploring network structure, dynamics, and function using NetworkX. in *Python in Science*, eds G. Varoquaux , T. Vaught, J. Millman (Pasadena, CA USA), pp 11–15.
20. A. Tversky, Features of similarity. *Psychological Review* **84**, 327–352 (1977).

21. S. A. Waraich, J. D. Victor, The Geometry of Low- and High-Level Perceptual Spaces. *The Journal of Neuroscience* **44** (2024).
22. M. N. Hebart, C. Y. Zheng, F. Pereira, C. I. Baker, Revealing the multidimensional mental representations of natural objects underlying human similarity judgements. *Nature Human Behaviour* **4**, 1173–1185 (2020).
23. P. Sinha, R. Russell, A Perceptually Based Comparison of Image Similarity Metrics. *Perception* **40**, 1269–1281 (2011).
24. R. Zhang, P. Isola, A. A. Efros, E. Shechtman, O. Wang (2018) The Unreasonable Effectiveness of Deep Features as a Perceptual Metric. in *IEEE/CVF Conference on Computer Vision and Pattern Recognition (CVPR)*, pp 586–595.
25. J. Zujovic, T. N. Pappas, D. L. Neuhoff, Structural Texture Similarity Metrics for Image Analysis and Retrieval. *IEEE T Image Process* **22**, 2545–2558 (2013).
26. S. Fu *et al.*, DreamSim: Learning New Dimensions of Human Visual Similarity using Synthetic Data. *Advances in Neural Information Processing Systems 36 (NeurIPS 2023)* **36** (2023).
27. S. E. Favila, H. Lee, B. A. Kuhl, Transforming the Concept of Memory Reactivation. *Trends Neurosci* **43**, 939–950 (2020).
28. A. X. Chang *et al.*, ShapeNet: An Information-Rich 3D Model Repository. http://dx.doi.org/10.48550/arXiv.1512.03012.
29. F. Wang, J. Mei, A. Yuille, "SCLIP: Rethinking Self-Attention for Dense Vision-Language Inference" in Computer Vision – ECCV 2024. (2025), 10.1007/978-3-031-72664-4_18 chap. Chapter 18, pp. 315–332.
30. R. Shinoda, K. Shiohara, "PetFace: A Large-Scale Dataset and Benchmark for Animal Identification" in Computer Vision – ECCV 2024. (2025), 10.1007/978-3-031-72649-1_2 chap. Chapter 2, pp. 19–36.
31. T. Valentine, Face-space models of face recognition. *Sci Psych S*, 83–113 (2001).
32. K. Dobs, J. Yuan, J. Martinez, N. Kanwisher, Behavioral signatures of face perception emerge in deep neural networks optimized for face recognition. *Proceedings of the National Academy of Sciences* **120** (2023).
33. M. Q. Hill *et al.*, Deep convolutional neural networks in the face of caricature. *Nature Machine Intelligence* **1**, 522–529 (2019).
34. E. Tousi, M. Mur, The face inversion effect through the lens of deep neural networks. *Proceedings of the Royal Society B: Biological Sciences* **291** (2024).
35. V. E. Strehle, N. K. Bendiksen, A. J. O'Toole, Deep convolutional neural networks are sensitive to face configuration. *Journal of Vision* **24** (2024).
36. P. Leroy, A. Mastropietro, M. Nurisso, F. Vaccarino (2025) Attributes shape the embedding space of face recognition models. in *International Conference on Machine Learning (ICML)* (Proceedings of Machine Learning Research), pp 33960–33983.
37. J. Križaj, R. O. Plesh, M. Banavar, S. Schuckers, V. Štruc, Deep Face Decoder: Towards understanding the embedding space of convolutional networks through visual reconstruction of deep face templates. *Engineering Applications of Artificial Intelligence* **132** (2024).
38. C. Summers, M. J. Dinneen, Nondeterminism and Instability in Neural Network Optimization. *International Conference on Machine Learning, Vol 139* **139** (2021).
39. J. L. Yuan, H; Ding, X; Xie, W; Li, Y; Zhao, W; Wan, K; Shi, J; Hu, X; Liu, Z (2025) Understanding and Mitigating Numerical Sources of Nondeterminism in LLM Inference. in *Advances in Neural Information Processing Systems (NeurIPS 2025)*.

40. P. Madhyastha, R. Jain, On Model Stability as a Function of Random Seed. *Proceedings of the 23rd Conference on Computational Natural Language Learning, Conll*, 929–939 (2019).
41. NVIDIA Corporation (2026) CUDA Programming Guide. pp 5–7, 69–70.
42. S. W. Chetlur, C; Vandermersch, P; Cohen, J; Tran, J; Catanzaro, B; Shelhamer, E, cuDNN: Efficient Primitives for Deep Learning. http://dx.doi.org/10.48550/arXiv.1410.0759.
43. S. Shanmugavelu *et al.*, Impacts of floating-point non-associativity on reproducibility for HPC and deep learning applications. *Proceedings of Sc24-W: Workshops of the International Conference for High Performance Computing, Networking, Storage and Analysis* 10.1109/Scw63240.2024.00028, 170–179 (2024).
44. K. M. He, X. Y. Zhang, S. Q. Ren, J. Sun (2016) Deep Residual Learning for Image Recognition. in *IEEE/CVF Conference on Computer Vision and Pattern Recognition (CVPR)*, pp 770–778.